\documentclass[journal,twoside,web]{ieeecolor}
\vfuzz=\maxdimen
\hfuzz=\maxdimen
\usepackage{tmi}
\usepackage{cite}
\usepackage{amsmath,amssymb,amsfonts}
\usepackage{multirow}
\usepackage{algorithmic}
\usepackage{graphicx}
\usepackage{textcomp}
\usepackage{booktabs}
\usepackage{color}
\usepackage{colortbl}
\definecolor{gray}{RGB}{192, 192, 192}
\definecolor{darkkcyan}{RGB}{0,138,218}
\usepackage[colorlinks=true, allcolors=darkkcyan,urlcolor = black]{hyperref}

\usepackage[switch]{lineno}

\def\BibTeX{{\rm B\kern-.05em{\sc i\kern-.025em b}\kern-.08em
    T\kern-.1667em\lower.7ex\hbox{E}\kern-.125emX}}
\begin{document}
\title{Nucleus-aware Self-supervised Pretraining Using Unpaired Image-to-image Translation for Histopathology Images}
\author{Zhiyun Song, Penghui Du, Junpeng Yan, Kailu Li, Jianzhong Shou, Maode Lai, Yubo Fan, and Yan Xu
\thanks{This work is supported by the National Natural Science Foundation in China under Grant 62022010, the Beijing Natural Science Foundation-Haidian District Joint Fund in China under Grant L222032, the Beijing hope run special fund of cancer foundation of China under Grant LC2018L02, the Fundamental Research Funds for the Central Universities of China from the State Key Laboratory of Software Development Environment in Beihang University in China, the 111 Project in China under Grant B13003, the high performance computing (HPC) resources at Beihang University. \textit{(Corresponding author: Yan Xu.)} }
\thanks{Z. Song, P. Du, J. Yan, K. Li, Y. Fan, and Y. Xu are with the School of Biological Science and Medical Engineering, State Key Laboratory of Software Development Environment, Key Laboratory of Biomechanics and Mechanobiology of Ministry of Education, Research Institute of Beihang University in Shenzhen, Beijing Advanced Innovation Center for Biomedical Engineering, Beihang University, Beijing 100191, China (e-mail: zhiyunsung@gmail.com; dupenghui@buaa.edu.cn; junpeng\_yan@buaa.edu.cn; likailu@buaa.edu.cn; yubofan@buaa.edu.cn; xuyan04@gmail.com).}
\thanks{J. Shou is with the Department of Urology, National Cancer Center/National Clinical Research Center for Cancer/ Cancer Hospital, Chinese Academy of Medical Sciences and Peking Union Medical College, Chaoyang District, Beijing 100021, China (e-mail: shoujianzhong@cicams.ac.cn).}
\thanks{M. Lai is with the Department of Pathology School of Medicine, Zhejiang University, Zhejiang Provincial Key Lab Disease Proteomics and Alibaba-Zhejiang University Joint Research Center of Future Healthcare, Hangzhou 310053, China (e-mail: lmd@zju.edu.cn).}
}

\maketitle

\begin{abstract}
Self-supervised pretraining attempts to enhance model performance by obtaining effective features from unlabeled data, and has demonstrated its effectiveness in the field of histopathology images.
Despite its success, few works concentrate on the extraction of nucleus-level information, which is essential for pathologic analysis.
In this work, we propose a novel nucleus-aware self-supervised pretraining framework for histopathology images.
The framework aims to capture the nuclear morphology and distribution information through unpaired image-to-image translation between histopathology images and pseudo mask images.
The generation process is modulated by both conditional and stochastic style representations, ensuring the reality and diversity of the generated histopathology images for pretraining. 
Further, an instance segmentation guided strategy is employed to capture instance-level information.
The experiments on 7 datasets show that the proposed pretraining method outperforms supervised ones on Kather classification, multiple instance learning, and 5 dense-prediction tasks with the transfer learning protocol, and yields superior results than other self-supervised approaches on 8 semi-supervised tasks. Our project is publicly available at \href{https://github.com/zhiyuns/UNITPathSSL}{https://github.com/zhiyuns/UNITPathSSL}.
\end{abstract}


\section{Introduction}
\label{sec:introduction}
\IEEEPARstart{H}{istopathology} 
serves as a crucial element for the diagnosis, prognosis, 
and analysis of therapeutic response for nearly all cancer types discovered\cite{Gurcan2009Pathreview,ludwig2005biomarkers}. 
In the field of computer-aided pathologic diagnosis (CAPD), fully supervised deep models dominate various pathology-related tasks, including 
cancer classification\cite{yang2019ems}, nuclei segmentation\cite{GRAHAM2019hovernet}, and molecular subtype identification\cite{sirinukunwattana2021image}.
However, these models rely on extensive annotation, especially for dense-prediction tasks that require instance-level annotations.
Fortunately, it is easier to obtain unlabeled histopathology images, which are expected to be utilized correctly to reduce the annotation burden for professional annotators.

One of the most popular approaches meeting the criteria is self-supervised pretraining, which learns generalized representation from unlabeled data.
It can be roughly categorized into discriminative and generative ones.
Currently, discriminative self-supervised pretraining is dominated by contrastive learning methods such as SimCLR\cite{Chen2020SimCLR}, MoCo v2\cite{chen2020mocov2}, and SimSiam\cite{Chen2021SimSia}.
On the other hand, the most recent development in generative self-supervised pretraining is based on denoising autoencoders\cite{Vincent2008DenoisingAutoencoders}, which use an encoder-decoder architecture to reconstruct corrupted images\cite{he2021mae, xie2021simmim, zhou2021ibot}.
While the corrupting procedure might work well for natural images that often contain large foreground objects against varying backgrounds, it is not ideal for histopathology images.
Simply corrupting histopathology images results in the loss of nuclei, which are often small but crucial for cancer identification, grading, and prognosis\cite{Gurcan2009Pathreview}.
For CAPD-related tasks, especially dense-prediction tasks, the importance of nuclei can also hardly be overstated\cite{GRAHAM2019hovernet, Koohbanani2021SelfPath, Mahmood2019GANseg2, Liu2020GANDA2, Gong2021StyleGenSeg, graham2021conic}.
Therefore, it is important to be aware of these small instances when tailoring self-supervised pretraining to histopathology images.

\begin{figure}[t]
    \centering
    \includegraphics[width=\linewidth]{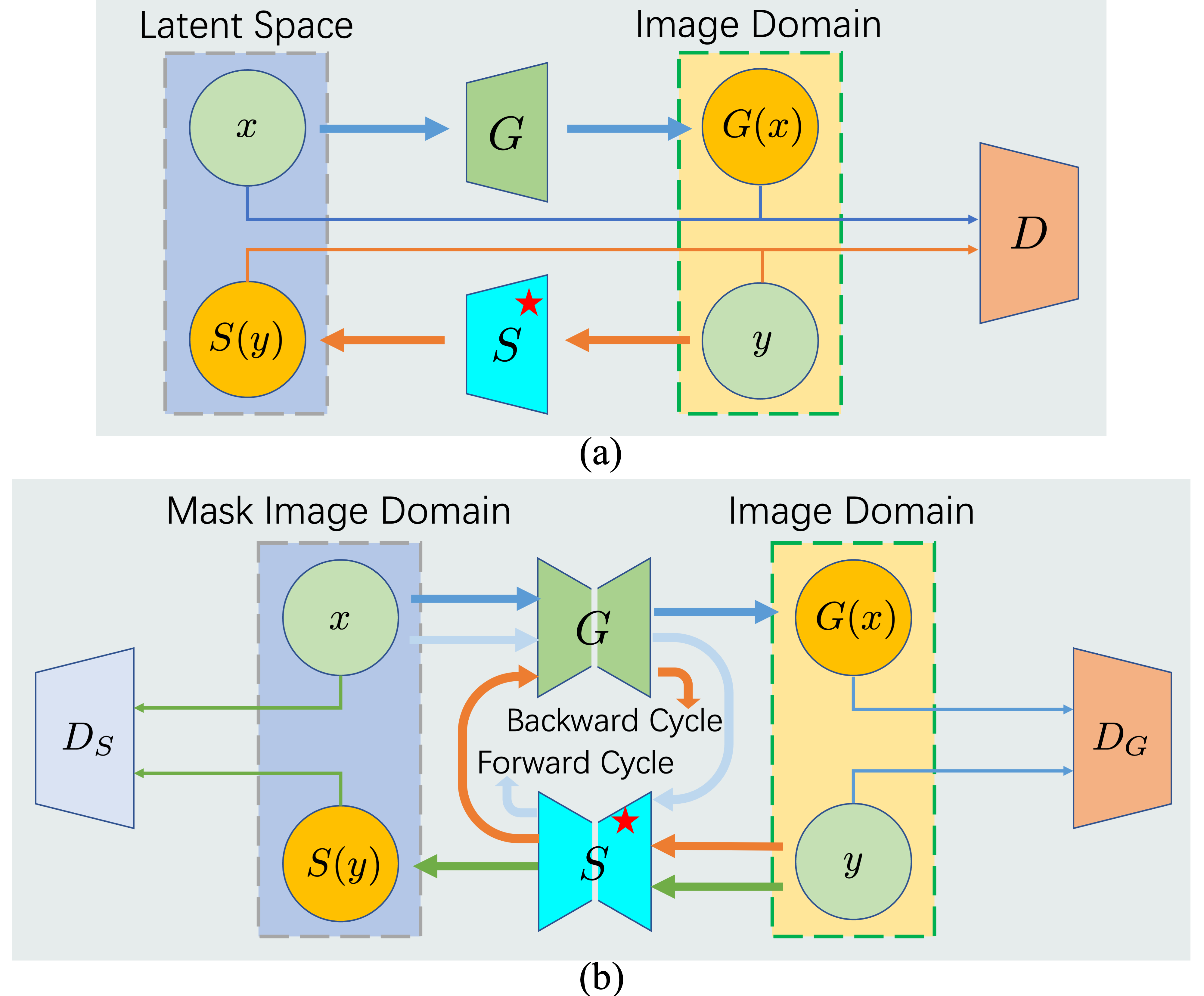}
	\caption{(a) Bidirectional pretraining framework in BiGAN. (b) Our CycleGAN-based pretraining framework. The original latent space is replace by the mask image domain, and the discriminators are designed to only distinguish images in each domain. The framework is further constrained by cycle consistency. Also, the generators $S$ and $G$ are both designed with encoder-decoder architectures instead of only the encoders. $\star$ means the network is used for downstream tasks.}
	\label{framework_comparison}
\end{figure}

As an extension of generative pretraining, adversarial pretraining with Generative Adversarial Networks (GAN) offers the possibility of nucleus-aware pretraining. 
It not only avoids the corrupting procedure, but also learns the data distribution prior for downstream tasks.
Most of the relevant works utilize a bidirectional framework that extends GAN with an extra encoder\cite{dumoulin2016ALI, donahue2016BiGAN, Donahue2019BiBigGAN}, as shown in \autoref{framework_comparison} \textcolor{darkkcyan}{(a)}.
However, the latent space of GAN is too abstract and redundant for fine-grained representation learning, hindering their performance on dense-prediction tasks.
Given that GAN can be employed as a semantic segmenter\cite{zhu2019GANseg1, Mahmood2019GANseg2},
whose procedure can also be viewed as image-to-image translation from images to their corresponding masks, the original latent space can be replaced with more specific representation, \emph{i.e.}, nucleus mask images.
This design is possible for histopathology images because of the low-cost feasibility of synthesizing mask images \cite{Mahmood2019GANseg2}.
In this way, the framework transforms naturally into unpaired image-to-image translation (UNIT) and is expected to capture nucleus-level information.
As illustrated in \autoref{framework_comparison} \textcolor{darkkcyan}{(b)}, CycleGAN\cite{Zhu2017CycleGAN} serves as a simple yet effective UNIT framework to implement the design.
It also consists of two networks that learn in an inverse manner, similar to the bidirectional framework.
The main difference lies in the cycle consistency, which regulates the translation and directs the network to capture semantic information.
Moreover, both generators are designed with an encoder-decoder architecture, which allows initializing the decoder for dense-prediction tasks.

It has been explored that the performance of adversarial pretraining is closely related to the quality of generated images\cite{Donahue2019BiBigGAN}.
Therefore, it is hard to yield optimal performance if we only pretrain with vanilla CycleGAN whose stochasticity is highly limited.
To overcome this challenge, a solution is to modulate the synthesis process with both conditional and stochastic style representations, which ensure the conditional correspondence and the intra-conditioning variety.
This co-modulation design was first proposed by CoModGAN\cite{zhao2021comodgan}, the most advanced large-scale image completion network.
By incorporating this design into the adversarial pretraining framework, the performance is expected to be significantly improved due to the enhanced quality of generated images. 

It is noteworthy that another generator $S$ is taken as the pretrained model.
Although cycle consistency is employed to enforce conditional constraints, it may not guarantee perfect matching between mask images and generated images.
Similar to previous works that enhanced the generation process with the guidance of segmentation\cite{Bazazian2022SegGuided1, Aakerberg2022SegGuided2, Ardino2021SegGuided3}, it is advantageous to incorporate the segmentation task into the pretraining framework.
However, relying solely on semantic segmentation may not provide effective information because the translation from histopathology images to mask images can be viewed as a form of semantic segmentation, as discussed previously.
Noticing the fact that we can conveniently obtain instance-level ground truth from mask images,
it is possible to augment the pretraining framework with instance segmentation, which can lead to a more accurate interpretation of instance objects.
Moreover, unlike many existing works which only pretrain the encoder, implementing instance segmentation in the pretraining process can yield a stronger initialization for downstream dense-prediction tasks.

Based on the analysis, we propose a novel nucleus-aware self-supervised
pretraining framework for histopathology images using a CycleGAN-based UNIT between the mask domain and the histopathology domain.
Pseudo mask images are first synthesized as the mask domain containing rich semantic information.
We integrate co-modulated generator into our framework to get realistic and diverse images for pretraining.
We also guide the framework with an instance segmentation task, which makes the constraints stricter and puts more focus on instance objects.

The novelties and contributions of this paper are summarized as follows:
\begin{itemize}
    \item We propose a novel adversarial self-supervised pretraining method within the framework of unpaired image-to-image translation, which is aware of nuclear instances in histopathology images.
    \item Co-modulation of both conditional and stochastic style representations is introduced for high-quality histopathology image generation conditioned on pseudo mask images, ensuring that the generated images provided for the pretrained network are realistic and diverse enough.
    \item We couple the framework with an instance segmentation task that offers instance-level representation, and is more effective for nucleus-aware pertaining than the common semantic segmentation guidance.
    \item The results for 7 transfer learning experiments and 8 semi-supervised experiments demonstrate that our method provides more effective and robust initialization for 5 different networks on 7 datasets, compared with other pretraining methods.

\end{itemize}

\section{Related Work}
Our work is related to four categories: (1) Self-supervised pretraining for histopathology images, (2) self-supervised pretraining with GAN, (3) unpaired image-to-image translation (UNIT), and (4) segmentation-guided synthesis.

\subsection{Self-supervised Pretraining for Histopathology Images}
Self-supervised pretraining has been an established topic 
in natural images for years, but studies attended to 
histopathology images are limited,
especially in terms of generative pretraining.
Koohbanani \emph{et al.}\cite{Koohbanani2021SelfPath} designed 
three pathology-specific pretext tasks based on the multi-scale nature and the special staining property of digital pathology.
Yang \emph{et al.}\cite{Yang2022CSCO} tailored contrastive learning to histopathology with stain vector perturbation and combined it with a cross-staining prediction task.
Luo \emph{et al.}\cite{luo2022maepath} enhanced the encoder of MAE\cite{he2021mae} with a self-distillation scheme that used tokens from visible histopathology patches.
All of the above methods pretrained the encoder solely for classification tasks.
Pretraining approaches which are also suitable for pathology-specific dense-prediction tasks are still under-explored.
Generative pretraining is believed to provide low-level representation that is advantageous for dense-prediction tasks\cite{Liu2021SSLreview, chen2022context}.
However, the direct application of generative pretraining to histopathology is not as effective as it is for natural images.
In this paper, we tailor generative self-supervised pretraining to histopathology.
Unlike previous works which ignore nuclear morphology and distribution, 
we guide the network to capture cellular information with pseudo mask images.
Moreover, unlike most previous studies that only pretrain the encoder for classification tasks, we simultaneously initialize 
the encoder and decoder for dense-prediction tasks.

\subsection{Self-supervised Pretraining with GAN}
First proposed by Goodfellow \emph{et al.}\cite{Goodfellow2014GAN}, 
Generative Adversarial Networks (GAN) is one of the most popular 
generative learning frameworks.
Recent works modified the architecture of GAN to generate high-quality images\cite{brock2018BigGAN, Karras2019StyleGAN, Kupyn2019DeblurGANv2, Karras2020StyleGAN2},
with StyleGAN\cite{Karras2019StyleGAN, Karras2020StyleGAN2} achieving impressive results for unconditional image generation.
Furthermore, CoModGAN\cite{zhao2021comodgan} tailored StyleGAN to image-to-image translation, achieving leading results for image completion.

Because of the unsupervised property of adversarial training, GAN also provides the possibility for adversarial pretraining\cite{Liu2021SSLreview}.
Previous works have considered the generator to be an implicit autoencoder by incorporating an additional encoder that maps images to 
latent representation\cite{dumoulin2016ALI, donahue2016BiGAN, Donahue2019BiBigGAN}.
DiRA\cite{Haghighi2022DiRA} combined adversarial learning with discriminative and restorative learning,
effectively guiding the encoder to capture more informative aspects of medical images.
Tao \emph{et al.} \cite{Tao2020Rubik} embedded adversarial training to Rubik’s Cube restoration, which adopts volume-wise transformations for context permutation.
In this work, we also pretrain a generator to provide 
powerful visual representation with adversarial self-supervised learning. 
Different from previous methods, 
we not only guide the generator to learn representation from original data, 
but also help it capture semantic information from high-quality images generated by CoModGAN within the framework of UNIT.

\subsection{Unpaired Image-to-image Translation}
To learn the mapping between two domains without paired images, CycleGAN\cite{Zhu2017CycleGAN} proposed a novel cycle consistency loss that preserves the structural information between domains. 
The bidirectional constraint has been tailored to several tasks. 
For instance, Mondal \emph{et al.}\cite{mondal2019CycleSSL} proposed a semi-supervised segmentation method by learning a bidirectional mapping between unlabeled images and available ground truth masks with CycleGAN.
Hoffman \emph{et al.}\cite{Hoffman2018CyCADA} proposed an effective domain adaptation method by adapting pixel-level and feature-level representation using the cycle consistency loss and the task loss.
These studies highlight that the cycle consistency constraint leads the generator to be aware of the semantic meaningful structure when applying style transfer between domains.
In this paper, we also use the cycle consistency constraint to preserve structural information, while enhancing the generation quality using CoModGAN and the instance segmentation guided strategy.
These approaches are incorporated into self-supervised pretraining.
To the best of our knowledge, none have additionally explored UNIT in the field of self-supervised pretraining for digital pathology at all.

\subsection{Segmentation Guided Synthesis}
\label{mask generation}
Studies have shown that the segmentation-guided strategy (SG) improves the performance of GANs by imposing spatial limitations on generated images
\cite{Bazazian2022SegGuided1, Aakerberg2022SegGuided2, Ardino2021SegGuided3, Gong2021StyleGenSeg}.
Bazazian \emph{et al.}\cite{Bazazian2022SegGuided1} guided dual-domain
image synthesis with part-based segmentation.
Aakerberg \emph{et al.}\cite{Aakerberg2022SegGuided2} 
utilized an auxiliary segmentation task to help produce accurate
super-resolution results.
Ardino \emph{et al.}\cite{Ardino2021SegGuided3} leveraged the predicted
segmentation map to facilitate the inpainting process, 
thereby improving the generation quality of images in complex scenes.
Most of the studies utilized semantic segmentation to provide guidance.
However, the strategy might not perform well in our framework because the information provided by semantic segmentation is similar to that in image-to-mask translation.
Therefore, we incorporate instance segmentation to provide more efficient guidance.
The most relevant method for our purposes has been proposed by 
Gong \emph{et al.}\cite{Gong2021StyleGenSeg}, who fused an
extra instance segmentation model into the generation procedure
in an adversarial manner.
However, this approach required a memory-consuming extra segmentation network.
In this paper, we further improve the method by sharing the backbone of the generator with an instance segmentation network, thereby increasing the pretraining efficiency.

\begin{figure*}[t]
	\centering
	\includegraphics[width=\textwidth]{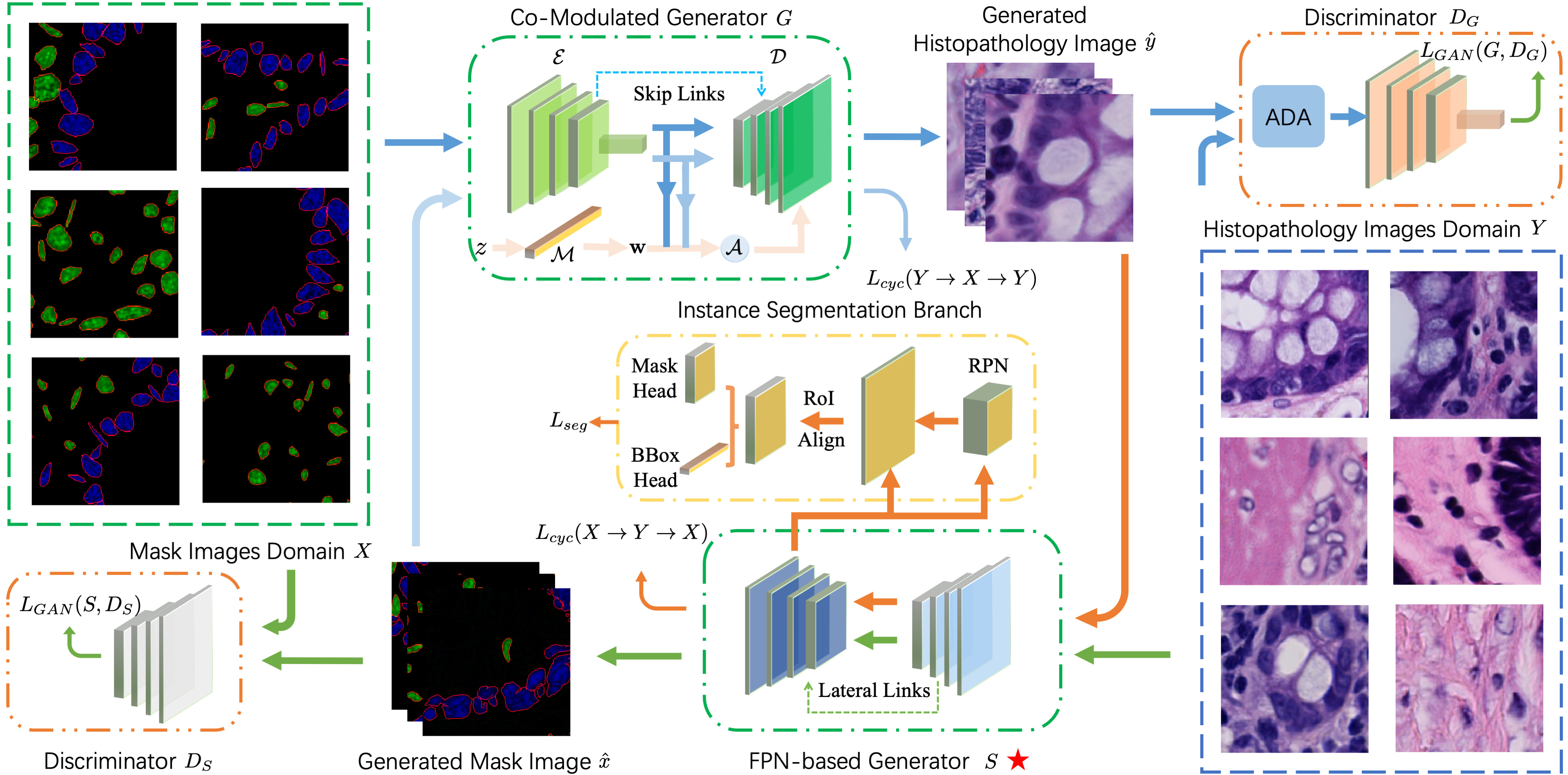}
\caption{Framework of our pretraining approach. The main idea is to pretrain the generator $S$ with unpaired image-to-image translation (UNIT) between mask image domain $X$ and histopathology image domain $Y$. $G$ produces the high-quality histopathology image $\hat{y}$ with the co-modulation of the conditional input $x$ and the style representation $\mathbf{w}$. Adaptive discriminator augmentation (ADA) is introduced to the discriminator $D_G$, which ensures the reality of $\hat{y}$ with $L_{GAN}(G, D_G)$. $S$ translates the histopathology image to mask image $\hat{x}$ with the constraint of $L_{GAN}(S, D_S)$. Cycle consistency loss $L_{cyc}$ is introduced to preserve the image structure during translation, with $G$ recovering original input $y$ for $\hat{x}$ and $S$ recovering original input $x$ for $\hat{y}$. An auxiliary instance segmentation branch is built upon the FPN-based architecture of $S$, with segmentation-guided strategy implemented during the translation from $\hat{y}$ to $x$ with $L_{seg}$. $\star$ means the corresponding network is used for downstream tasks.}
	\label{framework}
\end{figure*}

\section{Methods}
Our goal is to develop a novel nucleus-aware self-supervised pretraining method for histopathology images.
The framework learns the mapping between the mask image domain $X$ and histopathology image domain $Y$ with the CycleGAN-based UNIT, as shown in \autoref{framework}.
Generator $G$ produces histopathology image $\hat{y}$ conditioned by mask image $x$.
In order to generate images that are diverse enough for sufficient pretraining, we co-modulate $G$ by both the mask image $x$ and style representation $\mathbf{w}$.
Generator $S$ learns the reverse mapping from $Y$ to $X$, extracting semantic representation from $\hat{y}$ and $y$.
An auxiliary instance segmentation branch is implemented to provide instance-level information.

After self-supervised pretraining, we selectively take parts of $S$ to initialize the network for downstream tasks.
For classification tasks, we take the encoder of $S$ as the pretrained model.
For dense-prediction tasks, we are able to use the pretrained encoder, decoder, and even the segmentation branch to provide more comprehensive initialization for the segmenter, as we will detail in our experiments.

\subsection{Self-supervised Pretraining with UNIT}
We propose to capture the semantic information contained in histopathology images in an adversarial manner.
Similar to BiGAN, CycleGAN can also be regarded as a generalized bidirectional framework, as shown in \autoref{framework_comparison}.
The generators $G$ and $S$ respectively translate mask images and histopathology images to synthesized histopathology images and mask images.
Discriminators $D_G$ and $D_S$ learn to distinguish real and synthesized images generated from $G$ and $S$ with an adversarial loss $L_{GAN}$. 
For $G:X\rightarrow Y$, the objective $L_{GAN}(G, D_G)$ is defined as a non-saturation loss with $R_1$ regularization\cite{Karras2019StyleGAN}, which can be expressed respectively for the discriminator and the generator as:
\begin{equation}
\begin{aligned}
\min _{D_G} V(D_G) &= -\mathbb{E}_{y \sim p_{data}(y)}\left[\log D_{G}(y)\right] \\
& - \mathbb{E}_{x \sim p_{data}(x)}\left[\log(1-D_{G}(G(x)))\right] \\
& +\frac{{\gamma}_{g}}{2} \mathbb{E}_{y \sim p_{data}(y)} \left[\left\|\nabla D_{G}(y)\right\|^{2}\right] \\
\min _{G} V(G) &= -\mathbb{E}_{x \sim p_{data}(x)}\left[\log D_{G}(G(x))\right].
\end{aligned}
\label{nonsaturation_loss}\end{equation}

For $S:Y\rightarrow X$, least square loss\cite{mao2016cycleloss} with $R_1$ regularization is adopted as $L_{GAN}(S, D_S)$, whose objective functions are:
\begin{equation}
\begin{aligned}
\min _{D_S} V(D_S) &=\frac{1}{2} \mathbb{E}_{x \sim p_{data}(x)}\left[(D_{S}(x)-1)^{2}\right] \\
&+\frac{1}{2} \mathbb{E}_{y \sim p_{data}(y)}\left[(D_{S}(S(y)))^{2}\right] \\
&+\frac{{\gamma}_{s}}{2} \mathbb{E}_{x \sim p_{data}(x)} \left[\left\|\nabla D_{S}(x)\right\|^{2}\right]\\
\min _{S} V(S) &=\frac{1}{2} \mathbb{E}_{y \sim p_{data}(y)}\left[(D_{S}(S(y))-1)^{2}\right].
\end{aligned}
\label{leastsquare_loss_d}\end{equation}

The adversarial loss helps the generators produce realistic images which match the distribution of real data. 
As a result, $S$ is able to capture the nuclei distribution prior contained in mask images, yielding more reasonable nucleus-aware pretraining results.

Cycle consistency constraints are implemented to maintain structural information during the translation.
The main idea is to ensure that one of the generators reconstructs the original inputs given samples generated by another generator, which can also be denoted as $S(G(x))\approx x$ (the forward cycle consistency) and $G(S(y))\approx y$ (the backward cycle consistency).
The L1 norm is used to ensure the constraint, whose loss can be expressed as:
\begin{equation}
\begin{aligned}
L_{cyc}(G, S)
&=\mathbb{E}_{x \sim p_{data}}(x)\left[\|S(G(x))-x\|_{1}\right] \\
&+\mathbb{E}_{y \sim p_{data}}(y)\left[\|G(S(y))-y\|_{1}\right].
\end{aligned}
\label{cycle_consistency}\end{equation}

With the constraint of forward cycle consistency, \emph{i.e.} $S(G(x))\approx x$, generator $S$ learns to extract semantic features from high-quality histopathology images generated by $G$. 
Specifically, the network learns to identify nuclei, recognize their boundaries, and distinguish the epithelial ones. 
Combining the adversarial loss and the cycle consistency loss together, 
our framework learns semantic representation from both real and synthetic histopathology images.

\subsection{Unpaired Data Preparation}
The unsupervised translation between domain $X$ and domain $Y$ requires a substantial number of images in each domain.
We can easily acquire sufficient histopathology images for domain $Y$ by cropping patches from the scanned WSIs.
Due to the inability to obtain sufficient real nuclei masks for domain $X$, we arrange random polygons, representing nuclei, on a grid in order to generate arbitrary numbers of mask images.
The preparation of mask images is similar to that described in \cite{Mahmood2019GANseg2}, but we add more details to match the distribution of the nuclei and provide more information for high-quality generation.
The preparation of mask images can be roughly split into two parts: How the nuclei are distributed and how they are stylized.

\subsubsection{Distribution}
Recognizing the glandular structures is beneficial for the diagnosis and grading of carcinomas originating in several organs, including the prostate, breast, and colon\cite{Sirinukunwattana2015Glas}.
In these organs, nuclei are specially distributed due to the existence of glands. 
We simulate the phenomenon by positioning some of the nuclei around the lumens, which are represented by deformed, empty, and overlapping ovals.
To be specified, we randomly determine the center point, orientation angle, and length of the axis for each oval.
A polar coordinate system is then built upon the oval center, with nucleus centers being evenly distributed around the oval at each polar angle. 
The radial distance of each nucleus is perturbed, and more nuclei are randomly added at each polar angle.

\subsubsection{Stylization}
We stylize the masks of nuclei and place them on the centers determined by nuclei distribution.
Nuclei masks are generated as random polygons smoothed with Bézier interpolation, and we distinguish epithelial cells which form the glandular structures and other cells by placing their nuclei in different channels.
The intensity values of nuclei are perturbed to add stochasticity, and the nuclei boundaries are drawn in a separate channel in order to guide the network to focus more on boundary recognition.
Moreover, we allow some of the nuclei to slightly overlap with each other, which is a common phenomenon in histopathology.
Examples of the prepared mask images are shown in \autoref{framework}.

\begin{figure}[t]
\centering
\includegraphics[width=\linewidth]{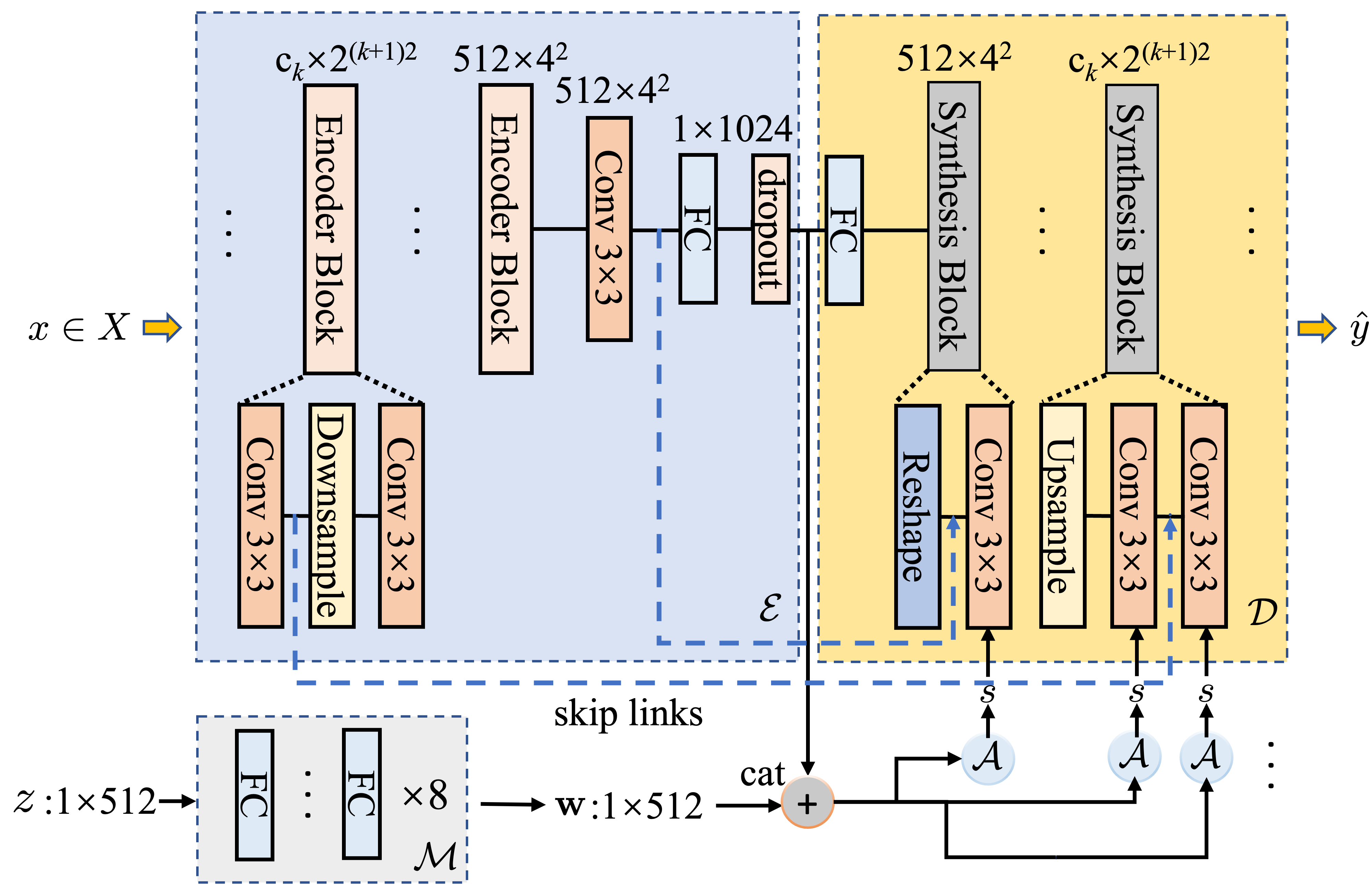}
\caption{Illustration of co-modulated generator $G$. The encoding network $\mathcal{E}$ (top left) takes a mask image $x$ as the input and generates the conditional style representation $\mathcal{E}(x)$ with a series of encoder blocks and a fully connected layer. 
The mapping network $\mathcal{M}$ (bottom left) transforms the noise vector $z$ into the stochastic style representation $w$.
The two representations are concatenated and produce the style vector $s$ with affine transformations before modulating the synthesis network $\mathcal{D}$ (top right).
For the first synthesis block of $\mathcal{D}$, the input is replaced by a feature map reshaped from the transformed conditional style representation.
The encoder blocks and synthesis blocks are linked with skip connections.
}
\label{CoModGAN}
\end{figure}

\subsection{Co-Modulated GAN}
The performance of our pretraining framework is directly related to the quality of generated histopathology images.
Recent works advance the generation quality of GANs with style-based modulation\cite{Karras2019StyleGAN, Karras2020StyleGAN2},
but the unconditional property of these GANs makes it hard to constrain the contents.
To bridge the gap between the image-conditioned and the style-based generator, one of the solutions is to modulate the generation process with both the stochastic style and conditional representations.
The strategy has been explored in CoModGAN\cite{zhao2021comodgan} and is proven to generate high-quality and diverse images conditioned by limited information.
Following this, we utilize the co-modulation strategy to improve the quality and stochasticity of generated histopathology images.
Moreover, we extend CoModGAN by adopting co-modulated StyleGAN2-ADA\cite{Karras2020Styleganada} to stabilize the pretraining in the limited data regime.

The main architecture of the co-modulated generator is shown in \autoref{CoModGAN}.
Given a mask image $x\in X$, $\mathcal{E}$ encodes it to a $4 \times 4$ feature map, which is flattened into conditional style representation $\mathcal{E} (x)$ with the dropout rate of 0.5. 
In addition, a mapping network $\mathcal{M}$ produces the stochastic style representation $\mathbf{w}$ by transforming a noise vector $z\sim N(0,\mathbf{I} )$.
The two representations are then concatenated and produce a style vector $s$ with the affine transform:
\begin{equation}
s = A(\mathcal{E} (x), \mathcal{M} (z)).
\label{comodgan}\end{equation}

The style vector $s$ is then used to modulate the synthesis network $\mathcal{D}$ via the "demodulation" operation applied to the convolution weight, as described in StyleGAN2.
Note that the original constant input of the first synthesis blocks is replaced by a $4 \times 4$ feature map, which is reshaped from the output of the fully connected layer following $\mathcal{E}$.
To further preserve the structure of mask images, skipping residual connections are implemented between $\mathcal{E}$ and $\mathcal{D}$ following CoModGAN \cite{zhao2021comodgan}.
$\mathcal{M}$ and $\mathcal{D}$ are designed to be the same as those in StyleGAN2, and the architecture of $\mathcal{E}$ is implemented the same with that in CoModGAN. 
We represent the whole generation as $\hat{y} = G(x)$ for simplification in the paper.

To further stabilize training, ADA\cite{Karras2020Styleganada} is adopted in the process of discriminating $\hat{y}$ and $y$.
Specifically, augmentations including pixel blitting, geometric transforms, and color transforms are applied for both $\hat{y}$ and $y$ before feeding them to the discriminator.
In order to avoid the leakage of the augmentations to the generator, 
the strength of conducting each augmentation is constrained by the scalar $p\in [0,1]$, which is determined by the degree of overfitting.
We use $r_t=\mathbb{E}[\mathrm{sign} (D_{\mathrm{train}})]$ as the overfitting heuristic,
and adjust $p$ once every four minibatchs:
\begin{equation}
p = \max\left\{ p + \frac{\mathrm{sign}(r_t-0.6) \times (N \times 4)} {500000} ,0\right\},
\label{adjustp}\end{equation}
where N is the batch size, and $p$ is initialized to be 0.
Then the augmentation is applied in sequence to the images with the probability of $p$.

The design of CoModGAN-ADA enhances the quality of generated histopathology images, and substantially improves the pretraining performance by providing more diverse synthesized images to $S$ in the forward cycle.

\subsection{Segmentation Guided Strategy}
\label{sub:sg}

We guide our pretraining framework with an auxiliary segmentation task, which has been proven to enhance the generation quality because of the extra attention to the region of interest\cite{Bazazian2022SegGuided1, Aakerberg2022SegGuided2, Ardino2021SegGuided3, Gong2021StyleGenSeg}.
In contrast to the semantic segmentation-guided strategy in most of the previous works, we utilize instance segmentation to provide the guidance.
Given a pseudo mask image $x$ generated by our algorithm, it is possible to obtain its instance-level label for nucleus segmentation.
In conjunction with the forward cycle consistency, we perform instance segmentation by adding an extra instance segmentation branch to $S$ so that it can also predict the instance-level label for $\hat{y}$, which should be identical to the label derived from $x$.

To make the generator be agnostic with different architectures of FPN-based instance segmentation networks, $S$ is designed to be similar to the generator of DeblurGAN-v2\cite{Kupyn2019DeblurGANv2}, which is based on Feature Pyramid Network (FPN) and works well with different backbones.
Five final feature maps in the bottom-up pathway of FPN are all up-sampled to the $\frac{1}{4}$ input size and summed up into one tensor.
The end of the network is designed the same as DeblurGAN-v2 to recover the original shape, except that batch normalization is replaced by instance normalization.

In order to conduct instance segmentation, we implement the region proposal network, bounding-box recognition head, and mask prediction head on top of the FPN structure of $S$.
The design of the instance segmentation branch is identical to those in Mask R-CNN\cite{He2017MRCNN}, whose loss is expressed as:
\begin{equation}
\begin{aligned}
L_{seg}(S)
&=L_{anchor-cls} + L_{anchor-reg} \\
&+ L_{bbox-cls} + L_{bbox-reg} + L_{mask}, 
\end{aligned}
\label{inst_seg}\end{equation}
in which $L_{anchor-cls}$ and $L_{anchor-reg}$ are the classification loss and bounding-box regression loss for the region proposal network,
 $L_{bbox-cls}$,$L_{bbox-reg}$, and $L_{mask}$ are the classification loss, bounding-box regression loss, and mask loss for the region of interest (RoI).
Note that the instance segmentation loss can be back-propagated to $G$ through $\hat{y}$.
Hence, the constraint augments the cycle consistency loss and guides $G$ to generate histopathology images that correspond more closely to their masks.
Moreover, the auxiliary task substantially improves the awareness of the nuclei in $\hat{y}$ and the ability to capture the semantic representation.

\subsection{Overall Losses and Training}
The proposed framework is composed of four parts: Generator $G$, $S$, and their corresponding discriminator $D_G$, $D_S$.
They are trained coherently so that:
 (1) $G$ takes pseudo mask images as conditional inputs and produces realistic translated images.
With adversarial learning, $D_G$ leads the generated images to match the data distribution of histopathology images;
 (2) $S$ tries to generate mask images conditioned on histopathology images with the guidance of $D_S$;
 (3) Structural information is preserved during the translation with cycle consistency constraints;
 and (4) $S$ is able to capture instance-level representation with an auxiliary segmentation task. 
 The overall losses corresponding to the above requirements can be written as:
\begin{equation}
\begin{aligned}
L
&=L_{GAN}(G, D_G) + \lambda _{1}L_{GAN}(S, D_S)  \\
&+ \lambda _{2}L_{cyc}(G, S) + \lambda _{3}L_{seg}(S),
\end{aligned}
\label{overall}\end{equation}
where the hyperparameters $\lambda _{1}$, $\lambda _{2}$, and $\lambda _{3}$ are used to balance different parts.
Note that instance segmentation is only introduced to the generator $S$ and is optimized with multi-task learning.

Due to the low generation quality of $G$ in the early training stage, we also design a two-stage training strategy to alleviate the mismatched convergence speed between $G$ and $S$.
After several iterations of training, when $S$ begins to overfit to generated images, $S$ and $D_S$ are reinitialized and another stage of training is performed.
Additional training stages will not improve the pretraining quality, as demonstrated in our subsequent experiment. 

\section{Experiments}
\subsection{Datasets}
\label{section:dataset}
Various datasets are included in our experiments for pretraining and for downstream tasks.
The basic information of each dataset is summarized in \autoref{tab:dataset_summary}.
\subsubsection{In-house}
We collect 1093 WSIs scanned from H\&E stained colorectal tissues in the First Affiliated Hospital of Zhejiang University.
Both normal and malignant slices are included and cropped into patches with a size of 512$\times $512 pixels.
Finally, a total of 160,000 unlabeled histopathology images at 40$\times$ objective magnification (0.25 $\mu $m/pixel) are prepared for pretraining.

\subsubsection{Kumar}
Kumar\cite{Kumar2017Monuseg} is a public nuclear instance segmentation dataset containing 30 annotated tiles extracted from 30 patients in 18 institutes.
These images are scanned and cropped from H\&E stained tissues at 40$\times$ objective magnification (0.25 $\mu $m/pixel) from 7 different organs (prostate, breast, colon, liver, kidney, stomach, and bladder) with the size of 1000$\times $1000.
According to the original work, 16 images are used for training and 14 for testing.
We only use the dataset for the downstream task in the transfer learning protocol due to the extremely small amounts of data. 

\subsubsection{Kather}
This dataset consists of colorectal histopathology images covering 9 classes for classification\cite{kather2019NCT}.
All images with the size of 224$\times $224 pixels at 20$\times$ objective magnification (0.5 $\mu $m/pixel) are obtained from the tissue bank of the National Center for Tumor diseases (NCT).
We only utilize two of the classes, namely the colorectal adenocarcinoma epithelium (TUM) and normal colon mucosa (NORM), resulting in 23,082 images for training and 1,976 images for testing.
Similar to \cite{Koohbanani2021SelfPath}, we randomly sample 20\% of the training data for validation.

\subsubsection{Lizard}
Lizard\cite{Graham2021Lizard} is the largest nuclear instance segmentation and classification dataset currently available. 
It consists of 6 subsets (GlaS, CRAG, CoNSeP, DPath, PanNuke, and TCGA) of colorectal histopathology images with an average of 1,016$\times$917 pixels at 20$\times$ objective magnification ($\sim $ 0.5$\mu $m/pixel).
Nearly half a million nuclei are boundary-labeled and classified into 6 subtypes (epithelial, neutrophil, lymphocyte, eosinophil, plasma, and connective tissue cells).
The subset TCGA is currently not available and is excluded from our study, resulting in a total of 238 images.
The two largest subsets (the Dpath subset with 69 images and the CRAG subset with 64 images) are used for downstream tasks.
We equally divide the two subsets into 3 parts for training, validating, and testing.

\subsubsection{Crag}
Crag \cite{GRAHAM2019crag} is a public dataset proposed for segmenting colorectal adenocarcinoma glands. 
It is composed of 213 H\&E images with a size of approximately 1,512$\times$1,516 pixels at 20$\times$ objective magnification ($\sim $ 0.5$\mu $m/pixel). We follow the official settings and divide the dataset into 173 images for training and 40 images for testing.

\subsubsection{Pannuke}
The PanNuke dataset \cite{gamper2019pannuke} comprises 7,901 tiles with 256$\times$256 pixels at either 20$\times$ or 40$\times$ magnification, obtained from over 20,000 whole slide images (WSIs) of 19 organs. The dataset includes annotations of 189,744 nuclei belonging to 5 different cell types, namely neoplastic, non-neoplastic epithelial, inflammatory, connective, and dead cells. Following the official settings, the dataset is divided into a training set, a validation set, and a test set.

\subsubsection{In-house-MIL}
We also collect 260 images with the size of 10,000 $\times$ 10,000 pixels at 40$\times$ magnification from the same source with the In-house dataset.
The dataset is composed of 130 malignant images and 130 normal images, which are annotated by an expert pathologist.
A 65\%-15\%-20\% split was used to split data for training, validation, and testing.
The dataset is used for multiple instance learning.
A bag indicates a set of patches with the size of 512$\times$512 extracted from an image
Positive bags are malignant images that contain cancerous cells, and negative bags are images that only contain normal cells.

\begin{table}[t]
\centering
\caption{\label{tab:dataset_summary}
Summary of the datasets used in our experiments. Mag. is the abbreviation for objective magnification, TL is the abbreviation for transfer learning, SSL is the abbreviation for semi-supervised learning.}
\resizebox{\columnwidth}{!}{%
\begin{tabular}{lcccccccc}
\hline
\multicolumn{1}{c|}{\multirow{2}{*}{Datasets}} & \multirow{2}{*}{Size} & \multirow{2}{*}{Mag.} & \multicolumn{3}{c}{Num of Images} & \multirow{2}{*}{Nuclei} & \multicolumn{2}{c}{Protocol} \\
\multicolumn{1}{l|}{} &  &  & Train & Val & Test &  & TL & SSL \\ \hline
\multicolumn{9}{c}{Datasets for Pretraining} \\ \hline
\multicolumn{1}{l|}{In-house} & 512$\times$512 & 40$\times$ & 160,000 & - & - & - & \checkmark &  \\
\multicolumn{1}{l|}{Lizard} & $\sim $1,055$\times$934 & 20$\times$ & 194 & - & - & - &  & \checkmark \\
\multicolumn{1}{l|}{Kather} & 224$\times$224 & 20$\times$ & 23,082 & - & - & - &  & \checkmark \\ \hline
\multicolumn{9}{c}{Datasets for downstream Tasks} \\ \hline
\multicolumn{1}{l|}{Kumar} & 1,000$\times$1,000 & 40$\times$ & 16 & - & 14 & 16,954 & \checkmark &  \\
\multicolumn{1}{l|}{Lizard-Dpath} & $\sim $1,255$\times$1,042 & 20$\times$ & 23 & 23 & 23 & 168,510 & \checkmark & \checkmark \\
\multicolumn{1}{l|}{Lizard-CRAG} & $\sim $1,503$\times$1,516 & 20$\times$ & 22 & 21 & 21 & 189,043 & \checkmark & \checkmark \\
\multicolumn{1}{l|}{Kather} & 224$\times$224 & 20$\times$ & 18,638 & 4,444 & 1,976 & - & \checkmark & \checkmark \\
\multicolumn{1}{l|}{Crag} & $\sim $1,512$\times$1,516 & 20$\times$ & 173 & - & 40 & - & \checkmark &  \\
\multicolumn{1}{l|}{Pannuke} & $ $256$\times$256 & 20$\times /$40$\times$ & 2,656 & 2,523 & 2,722 & 189,744 & \checkmark &  \\
\multicolumn{1}{l|}{In-house-MIL} & $ $10,000$\times$10,000 & 40$\times$ & 169 & 39 & 52 & - & \checkmark &  \\
\hline
\end{tabular}%
}
\end{table}

\subsection{Evaluation}
\subsubsection{Evaluation Protocols}
The pretraining quality is accessed using the transfer learning (TL) and semi-supervised learning (SSL) protocol.
For TL, the framework is pretrained from scratch on the in-house dataset, followed by supervised training on various downstream tasks with end-to-end fine-tuning.
For SSL, models are pretrained on all training data and fine-tuned on parts of the dataset.

\subsubsection{Evaluation Metrics}
(1) Classification.
Accuracy (Acc) and F1 score are used for classification tasks, including patch-wise classification and multiple instance learning.
(2) Nuclear Segmentation and Detection.
Aggregated Jaccard Index (AJI)\cite{Kumar2017Monuseg} is reported to evaluate the segmentation quality, and F1 score with an IoU threshold of 0.5 is used to evaluate the detection quality.
(3) Multi-class Nuclear Segmentation and Detection.
If the further classification of nuclei is required, the segmentation quality is evaluated with multi-class panoptic quality (mPQ+) proposed in the CoNIC challenge\cite{graham2021conic}.
The metric is the average across per-class PQ, whose statics are calculated over all images.
F1 score averaged across all classes is reported to quantify the detection quality.
(4) Gland Segmentation.
We report object-level Dice and Hausdorff distance, which is used in the GlaS challenge\cite{Sirinukunwattana2017glas} to evaluate the instance-level gland segmentation quality.
We run each downstream task five times with different random seeds and report the
mean, standard deviation, and statistical analysis based
on independent two-sample t-test.

\subsection{Experiment Setting}
\subsubsection{Pretraining}
ResNet-50 is used as the feature extractor in the bottom-up pathway of $S$.
The normalization layers in the FPN and the segmentation branch are implemented with Group Normalization (GN) instead of Batch Normalization (BN) to alleviate the problem caused by the small batch size\cite{He2019RethinkGN}.
$D_G$ and $D_S$ maintain the design in StyleGAN2-ADA and CycleGAN, respectively.
We rescale the input images to be 40$\times$ in order so that networks can concentrate on fine-grained details.
Images are then resized in a range of 0.8$\times \sim$ 1.0$\times$ of original size and cropped to 256$\times$256 before feeding to our framework.
We optimize the framework with batch size 12 and Adam optimizer with $\beta 1 = 0$, $\beta2 = 0.99$, and $\epsilon = 10^{{}-8}$.
Hyperparameters for the instance segmentation branch are set following Mask-RCNN.
We also follow the settings of StyleGAN2 when training $G$, except that the lazy regularization and the path length regularization are disabled.
The weight-balancing hyperparameters $\lambda _{1}$, $\lambda _{2}$, and $\lambda _{3}$ are empirically set to 2.0, 10.0, and 2.0.
Both the $\gamma _{g}$ and $\gamma _{s}$ for $R_1$ regularization are set to be 1.0.
For the two-stage training strategy, we train the framework for 40k iterations with learning rate of $4.0 \times 10^{{}-4}$ in the first stage, and 25k iterations with learning rate of $1.0 \times 10^{{}-4}$ in the second stage.
The whole pretraining process is conducted on 4 GeForce RTX 2080 Ti GPUs.

\begin{table*}[ht]
\caption{\label{tab:transfer_experiment}
Performance comparison with the transfer learning protocol. * denotes supervised pretraining on an excluded dataset. For each metric, the result that ranks in the 1st place are shown in bold, and that ranks in the 2nd place are underlined. For the Kather dataset, we only use 1\% of training data (178 images) for fine-tuning.}
\resizebox{\textwidth}{!}{%
\begin{tabular}{lcccccccccccccc}
\hline
\multicolumn{1}{c}{\multirow{2}{*}{Method}} & \multicolumn{2}{c}{Kumar} & \multicolumn{2}{c}{Kather} & \multicolumn{2}{c}{Lizard-Dpath} & \multicolumn{2}{c}{Lizard-CRAG} & \multicolumn{2}{c}{PanNuke} & \multicolumn{2}{c}{Crag} & \multicolumn{2}{c}{In-house-MIL} \\
\multicolumn{1}{c}{} & AJI & F1 & Acc & F1 & mPQ+ & F1 & mPQ+ & F1 & mPQ+ & F1 & Obj Dice & Obj Haus & Acc & F1 \\ \hline
\multicolumn{15}{l}{\textbf{Baselines}} \\
Random & \begin{tabular}[c]{@{}c@{}}0.4815\\ ±0.0025\end{tabular} & \begin{tabular}[c]{@{}c@{}}0.6461\\ ±0.0060\end{tabular} & \begin{tabular}[c]{@{}c@{}}0.8603\\ ±0.0108\end{tabular} & \begin{tabular}[c]{@{}c@{}}0.8545\\ ±0.0110\end{tabular} & \begin{tabular}[c]{@{}c@{}}0.3288\\ ±0.0064\end{tabular} & \begin{tabular}[c]{@{}c@{}}0.3991\\ ±0.0083\end{tabular} & \begin{tabular}[c]{@{}c@{}}0.3468\\ ±0.0114\end{tabular} & \begin{tabular}[c]{@{}c@{}}0.4431\\ ±0.0121\end{tabular} & \begin{tabular}[c]{@{}c@{}}0.3915\\ ±0.0028\end{tabular} & \begin{tabular}[c]{@{}c@{}}0.4854\\ ±0.0033\end{tabular} & \begin{tabular}[c]{@{}c@{}}0.8011\\ ±0.0066\end{tabular} & \begin{tabular}[c]{@{}c@{}}248.26\\ ±10.33\end{tabular} & \begin{tabular}[c]{@{}c@{}}0.5192\\ ±0.0593\end{tabular} & \begin{tabular}[c]{@{}c@{}}0.6753\\ ±0.0259\end{tabular} \\
ImageNet* & \begin{tabular}[c]{@{}c@{}}0.5725\\ ±0.0043\end{tabular} & \begin{tabular}[c]{@{}c@{}}0.7466\\ ±0.0081\end{tabular} & \begin{tabular}[c]{@{}c@{}}\underline{0.9547}\\ ±0.0026\end{tabular} & \begin{tabular}[c]{@{}c@{}}\underline{0.9520}\\ ±0.0031\end{tabular} & \begin{tabular}[c]{@{}c@{}}0.3859\\ ±0.0069\end{tabular} & \begin{tabular}[c]{@{}c@{}}0.4766\\ ±0.0073\end{tabular} & \begin{tabular}[c]{@{}c@{}}0.3925\\ ±0.0091\end{tabular} & \begin{tabular}[c]{@{}c@{}}0.5201\\ ±0.0097\end{tabular} & \begin{tabular}[c]{@{}c@{}}0.4147\\ ±0.0022\end{tabular} & \begin{tabular}[c]{@{}c@{}}0.5364\\ ±0.0028\end{tabular} & \begin{tabular}[c]{@{}c@{}}0.8442\\ ±0.0051\end{tabular} & \begin{tabular}[c]{@{}c@{}}191.74\\ ±8.43\end{tabular} & \begin{tabular}[c]{@{}c@{}}0.7885\\ ±0.0193\end{tabular} & \begin{tabular}[c]{@{}c@{}}0.7317\\ ±0.0118\end{tabular} \\ \hline
\multicolumn{15}{l}{\textbf{Generalized Pretraining Approaches}} \\
MoCo v2\cite{chen2020mocov2} & \begin{tabular}[c]{@{}c@{}}\underline{0.5873}\\ ±0.0023\end{tabular} & \begin{tabular}[c]{@{}c@{}}0.7481\\ ±0.0058\end{tabular} & \begin{tabular}[c]{@{}c@{}}0.9478\\ ±0.0088\end{tabular} & \begin{tabular}[c]{@{}c@{}}0.9481\\ ±0.0089\end{tabular} & \begin{tabular}[c]{@{}c@{}}0.3883\\ ±0.0051\end{tabular} & \begin{tabular}[c]{@{}c@{}}\textbf{0.5007}\\ ±0.0068\end{tabular} & \begin{tabular}[c]{@{}c@{}}0.4282\\ ±0.0074\end{tabular} & \begin{tabular}[c]{@{}c@{}}0.5295\\ ±0.0088\end{tabular} & \begin{tabular}[c]{@{}c@{}}\underline{0.4379}\\ ±0.0043\end{tabular} & \begin{tabular}[c]{@{}c@{}}0.5468\\ ±0.0045\end{tabular} & \begin{tabular}[c]{@{}c@{}}0.8508\\ ±0.0038\end{tabular} & \begin{tabular}[c]{@{}c@{}}\underline{157.23}\\ ±6.49\end{tabular} & \begin{tabular}[c]{@{}c@{}}0.7152\\ ±0.0216\end{tabular} & \begin{tabular}[c]{@{}c@{}}0.7222\\ ±0.0204\end{tabular} \\
Simsiam\cite{Chen2021SimSia} & \begin{tabular}[c]{@{}c@{}}0.5517\\ ±0.0039\end{tabular} & \begin{tabular}[c]{@{}c@{}}0.7109\\ ±0.0079\end{tabular} & \begin{tabular}[c]{@{}c@{}}0.8977\\ ±0.0051\end{tabular} & \begin{tabular}[c]{@{}c@{}}0.8941\\ ±0.0050\end{tabular} & \begin{tabular}[c]{@{}c@{}}0.3467\\ ±0.0037\end{tabular} & \begin{tabular}[c]{@{}c@{}}0.4225\\ ±0.0048\end{tabular} & \begin{tabular}[c]{@{}c@{}}0.3769\\ ±0.0067\end{tabular} & \begin{tabular}[c]{@{}c@{}}0.4572\\ ±0.0079\end{tabular} & \begin{tabular}[c]{@{}c@{}}0.4230\\ ±0.0029\end{tabular} & \begin{tabular}[c]{@{}c@{}}0.5251\\ ±0.0038\end{tabular} & \begin{tabular}[c]{@{}c@{}}0.8508\\ ±0.0063\end{tabular} & \begin{tabular}[c]{@{}c@{}}159.54\\ ±9.37\end{tabular} & \begin{tabular}[c]{@{}c@{}}0.7793\\ ±0.0434\end{tabular} & \begin{tabular}[c]{@{}c@{}}0.7247\\ ±0.0399\end{tabular} \\ \hline
\multicolumn{15}{l}{\textbf{Pretraining Approaches for Medical Images}} \\
Mormont*\cite{Mormont2021SupPath} & \begin{tabular}[c]{@{}c@{}}0.5839\\ ±0.0068\end{tabular} & \begin{tabular}[c]{@{}c@{}}0.7470\\ ±0.0074\end{tabular} & \begin{tabular}[c]{@{}c@{}}0.9469\\ ±0.0035\end{tabular} & \begin{tabular}[c]{@{}c@{}}0.9433\\ ±0.0033\end{tabular} & \begin{tabular}[c]{@{}c@{}} 0.3926\\ ±0.0037\end{tabular} & \begin{tabular}[c]{@{}c@{}}0.4811\\ ±0.0049\end{tabular} & \begin{tabular}[c]{@{}c@{}}0.4324\\ ±0.0077\end{tabular} & \begin{tabular}[c]{@{}c@{}}0.5345\\ ±0.0082\end{tabular} & \begin{tabular}[c]{@{}c@{}}0.4369\\ ±0.0021\end{tabular} & \begin{tabular}[c]{@{}c@{}}\underline{0.5475}\\ ±0.0026\end{tabular} & \begin{tabular}[c]{@{}c@{}}0.8549\\ ±0.0044\end{tabular} & \begin{tabular}[c]{@{}c@{}}159.38\\ ±9.11\end{tabular} & \begin{tabular}[c]{@{}c@{}}\underline{0.8292}\\ ±0.0211\end{tabular} & \begin{tabular}[c]{@{}c@{}}\underline{0.8000}\\ ±0.0192\end{tabular} \\
DiRA\cite{Haghighi2022DiRA} & \begin{tabular}[c]{@{}c@{}}0.5850\\ ±0.0043\end{tabular} & \begin{tabular}[c]{@{}c@{}}\underline{0.7487}\\ ±0.0053\end{tabular} & \begin{tabular}[c]{@{}c@{}}0.9311\\ ±0.0046\end{tabular} & \begin{tabular}[c]{@{}c@{}}0.9279\\ ±0.0046\end{tabular} & \begin{tabular}[c]{@{}c@{}} \underline{0.3963}\\ ±0.0057\end{tabular} & \begin{tabular}[c]{@{}c@{}}0.4827\\ ±0.0072\end{tabular} & \begin{tabular}[c]{@{}c@{}}\underline{0.4362}\\ ±0055\end{tabular} & \begin{tabular}[c]{@{}c@{}}\underline{0.5431}\\ ±0.0064\end{tabular} & \begin{tabular}[c]{@{}c@{}}0.4375\\ ±0.0057\end{tabular} & \begin{tabular}[c]{@{}c@{}}0.5465\\ ±0.0066\end{tabular} & \begin{tabular}[c]{@{}c@{}}0.8491\\ ±0.0039\end{tabular} & \begin{tabular}[c]{@{}c@{}}157.51\\ ±6.73\end{tabular} & \begin{tabular}[c]{@{}c@{}}0.8016\\ ±0.0539\end{tabular} & \begin{tabular}[c]{@{}c@{}}0.7789\\ ±0.0331\end{tabular} \\
Ciga\cite{ciga2022contrastivepath} & \begin{tabular}[c]{@{}c@{}}0.5693\\ ±0.0089\end{tabular} & \begin{tabular}[c]{@{}c@{}}0.7380\\ ±0.0112\end{tabular} & \begin{tabular}[c]{@{}c@{}}0.9266\\ ±0.0074\end{tabular} & \begin{tabular}[c]{@{}c@{}}0.9230\\ ±0.0066\end{tabular} & \begin{tabular}[c]{@{}c@{}}0.3418\\ ±0.0057\end{tabular} & \begin{tabular}[c]{@{}c@{}}0.4171\\ ±0.0089\end{tabular} & \begin{tabular}[c]{@{}c@{}}0.3673\\ ±0.0079\end{tabular} & \begin{tabular}[c]{@{}c@{}}0.4501\\ ±0.0085\end{tabular} & \begin{tabular}[c]{@{}c@{}}0.3964\\ ±0.0038\end{tabular} & \begin{tabular}[c]{@{}c@{}}0.4956\\ ±0.0057\end{tabular} & \begin{tabular}[c]{@{}c@{}}\underline{0.8616}\\ ±0.0057\end{tabular} & \begin{tabular}[c]{@{}c@{}}168.14\\ ±9.12\end{tabular} & \begin{tabular}[c]{@{}c@{}}0.7582\\ ±0.0305\end{tabular} & \begin{tabular}[c]{@{}c@{}}0.7689\\ ±0.0256\end{tabular} \\
\rowcolor{gray} Ours & \begin{tabular}[c]{@{}c@{}}\textbf{0.5932}\\ ±0.0025\end{tabular} & \begin{tabular}[c]{@{}c@{}}\textbf{0.7591}\\ ±0.0044\end{tabular} & \begin{tabular}[c]{@{}c@{}}\textbf{0.9588}\\ ±0.0028\end{tabular} & \begin{tabular}[c]{@{}c@{}}\textbf{0.9560}\\ ±0.0029\end{tabular} & \begin{tabular}[c]{@{}c@{}} \textbf{0.3980}\\ ±0.0041\end{tabular} & \begin{tabular}[c]{@{}c@{}} \underline{0.4895}\\ ±0.0057\end{tabular} & \begin{tabular}[c]{@{}c@{}}\textbf{0.4377}\\ ±0.0056\end{tabular} & \begin{tabular}[c]{@{}c@{}}\textbf{0.5447}\\ ±0.0069\end{tabular} & \begin{tabular}[c]{@{}c@{}}\textbf{0.4432}\\ ±0.0022\end{tabular} & \begin{tabular}[c]{@{}c@{}}\textbf{0.5492}\\ ±0.0031\end{tabular} & \begin{tabular}[c]{@{}c@{}}\textbf{0.8649}\\ ±0.0025\end{tabular} & \begin{tabular}[c]{@{}c@{}}\textbf{154.15}\\ ±5.70\end{tabular} & \begin{tabular}[c]{@{}c@{}}\textbf{0.8462}\\ ±0.0479\end{tabular} & \begin{tabular}[c]{@{}c@{}}\textbf{0.8621}\\ ±0.0409\end{tabular} \\ \hline

\end{tabular} %
}
\end{table*}

\begin{table}[t]
\centering
\caption{\label{tab:compare_hovernet}
Comparison with state-of-the-art results for nuclear segmentation on Kumar. The results of UNet, Mask-RCNN, Micro-Net, CIA-Net, and Hover-Net are directly copied from Hover-Net, while the results of DSF-CNN and REU-Net are copied from their original articles.}
\begin{tabular}{lccc}
\hline
 & \multicolumn{3}{c}{Kumar} \\
 & Dice & AJI & PQ \\ \hline
\multicolumn{4}{l}{\textbf{Segmenters}} \\
UNet \cite{Ronneberger2015UNet} & 0.758 & 0.556 & 0.478 \\
Mask-RCNN \cite{He2017MRCNN} & 0.760 & 0.546 & 0.509 \\
Micro-Net \cite{RAZA2019micronet} & 0.797 & 0.560 & 0.519 \\
Panoptic FPN \cite{Kirillov2019PanopticFPN} & 0.805 & 0.573 & 0.557 \\
CIA-Net \cite{Zhou2019CiaNet} & 0.818 & 0.620 & 0.577 \\
Hover-Net \cite{GRAHAM2019hovernet} & 0.826 & 0.618 & 0.597 \\
DSF-CNN \cite{Graham2020dsfcnn} & 0.826 & - & 0.600 \\
REU-Net \cite{Jian2022REUNet} & 0.826 & 0.636 & 0.604 \\ \hline
\multicolumn{4}{l}{\textbf{Pretrained Hover-Net}} \\
Hover-Net (MoCo v2) & 0.842 & 0.640 & 0.612 \\
Hover-Net (Simsiam) & 0.841 & 0.628 & 0.602 \\
Hover-Net (Mormont) & 0.845 & 0.645 & 0.615 \\
Hover-Net (DiRA) & 0.836 & 0.624 & 0.586 \\
Hover-Net (Ciga) & 0.837 & 0.631 & 0.597 \\
\rowcolor{gray} Hover-Net (Ours) & \textbf{0.853} & \textbf{0.657} & \textbf{0.625} \\ \hline
\end{tabular}
\end{table}

\subsubsection{Fine-tuning}
For classification tasks on Kather, we initialize the ResNet backbone with our pretrained model and build a classifier head with adaptive average pooling, fully-connected layer, and softmax on top of the backbone.
For the nuclear detection and segmentation task on Kumar, we follow \cite{Liu2020GANDA2} and use Panoptic FPN\cite{Kirillov2019PanopticFPN} as the segmenter.
The weight of FPN is initialized by our pretraining approach, whereas the semantic segmentation branch and instance segmentation branch are initialized randomly.
For multi-class nuclear segmentation and detection on PanNuke and the subsets of Lizard, we utilize Mask-RCNN, whose FPN is initialized by our method.
For gland segmentation on Crag, we take UperNet\cite{Xiao2018upernet} as the semantic segmenter and initialize the ResNet-50 backbone.
In this task, both the objects and contours are predicted to help separate the touching instances following \cite{Graham2020dsfcnn}.
Multiple instance learning experiment is implemented following C2C \cite{sharma2021c2c}, whose backbone is replaced with ResNet-50.
The learning rate, batch size, and optimizer of each experiment are determined with grid search.

\begin{table}[t]
\centering
\caption{\label{tab:kumar_results}
Average AJI across 7 organ types on the Kumar dataset with the transfer learning protocol.}
\resizebox{\columnwidth}{!}{%
\begin{tabular}{lccccccc>{\columncolor{gray}}c}
\hline
\multicolumn{1}{c}{Organ} & Baseline & ImageNet & MoCo v2 & Simsiam & Mormont & DiRA & Ciga & Proposed \\
\hline
Stomach & 0.4821 & 0.5440 & 0.6590 & 0.6332 & \textbf{0.6678} & 0.6572 & 0.6317 & 0.6206 \\
Colon & 0.3933 & 0.4939 & 0.5001 & 0.4066 & 0.4545 & 0.4967 & 0.4813 & \textbf{0.5008} \\
Bladder & 0.5333 & 0.6364 & 0.5768 & 0.5652 & 0.5957 & 0.5847 & 0.5505 & \textbf{0.6404} \\
Prostate & 0.5360 & 0.6139 & 0.6304 & 0.5953 & 0.6234 & 0.6257 & 0.6280 & \textbf{0.6302} \\
Liver & 0.4405 & 0.5370 & \textbf{0.5455} & 0.4789 & 0.5253 & 0.5206 & 0.5190 & 0.5323 \\
Kidney & 0.5216 & 0.6025 & 0.6075 & 0.6176 & 0.6185 & 0.6145 & 0.6090 & \textbf{0.6226} \\
Breast & 0.4639 & 0.5795 & 0.5922 & 0.5652 & 0.6021 & 0.5960 & 0.5659 & \textbf{0.6056} \\
\hline
Overall & \begin{tabular}[c]{@{}c@{}}0.4815\\ ±0.0025\end{tabular} & \begin{tabular}[c]{@{}c@{}}0.5725\\ ±0.0043\end{tabular} & \begin{tabular}[c]{@{}c@{}}0.5873\\ ±0.0023\end{tabular} & \begin{tabular}[c]{@{}c@{}}0.5517\\ ±0.0039\end{tabular} & \begin{tabular}[c]{@{}c@{}}0.5839\\ ±0.0068\end{tabular} & \begin{tabular}[c]{@{}c@{}}0.5850\\ ±0.0043\end{tabular} & \begin{tabular}[c]{@{}c@{}}0.5693\\ ±0.0089\end{tabular} & \begin{tabular}[c]{@{}c@{}}\textbf{0.5932}\\ ±0.0025\end{tabular} \\
\hline
\end{tabular}
}
\end{table}

\subsection{Transfer Learning Experiments}
\label{transfer_experiment}
We experiment on our in-house dataset and prepare mask images with equivalent amounts of data for pretraining.
We compare our approach against the two most recent self-supervised pretraining methods for medical images (DiRA\cite{Haghighi2022DiRA} and Ciga-SimCLR\cite{ciga2022contrastivepath}), and two methods for natural images (MoCo v2\cite{chen2020mocov2} and Simsiam\cite{Chen2021SimSia}).
The primary hyperparameters of these pretraining methods have been tuned referring to the ablation study of their original work on Kumar (\emph{e.g.}, the momentum value and the softmax temperature of MoCo v2).
Supervised pretraining on ImageNet and Mormont \emph{et al.}\cite{Mormont2021SupPath} are also included for comparison.
Note that we can only initialize the ResNet backbone for downstream tasks with these approaches.
 
\subsubsection{Multi-organ Nuclear Detection and Segmentation}
As shown in \autoref{tab:transfer_experiment},
although no extra annotation is used in our pretraining framework, we outperform supervised pretraining on ImageNet (p$<$0.05) and Mormont \emph{et al.} (p$<$0.05) on Kumar.
Moreover, our method produces better results than previous SOTA pretraining approaches (p$<$0.05, compared with MoCo v2), indicating that our nucleus-aware method is more effective for the dense-prediction task and generalizes better to histopathology images from various organs.
The per-organ results reported in \autoref{tab:kumar_results} indicate that our method consistently benefits nuclear segmentation across various organs. We not only improve the performance on the organs with glandular structures (\emph{e.g.}, colon, prostate, and breast), but also benefit that without the special structure (\emph{e.g.}, bladder).
We also compare with the SOTA methods for nuclear segmentation on Kumar.
To this end, we use HoVer-Net\cite{GRAHAM2019hovernet} as the segmenter, and replace the Preact-ResNet-50 backbone with the pretrained ResNet-50, similar to the previous work \cite{mohta2022mrl}.
As reported in \autoref{tab:compare_hovernet}, it is interesting to find that all the pretraining methods improve the performance of Hover-Net (Dice), among which we achieve new SOTA results on Kumar.

\begin{table}[t]
\centering
\caption{\label{tab:pannuke_results}
Average mPQ+ across 19 tissue types on the Pannuke dataset with the transfer learning protocol.}
\resizebox{\columnwidth}{!}{%
\begin{tabular}{lccccccc>{\columncolor{gray}}c}
\hline
\multicolumn{1}{c}{Organ} & Baseline & ImageNet & MoCo v2 & Simsiam & Mormont & DiRA & Ciga & Proposed \\
\hline
Adrenal & 0.3303 & 0.3698 & 0.3835 & 0.3807 & \textbf{0.3854} & 0.3774 & 0.3561 & 0.3807 \\
Bile Duct & 0.3328 & \textbf{0.3696} & 0.3560 & 0.3510 & 0.3630 & 0.3648 & 0.3161 & 0.3613 \\
Bladder & 0.3170 & 0.2918 & 0.3330 & 0.3397 & 0.3221 & 0.3161 & 0.3245 & \textbf{0.3566} \\
Breast & 0.3773 & 0.3890 & 0.4022 & 0.4040 & 0.3956 & 0.4001 & 0.3690 & \textbf{0.4190} \\
Cervix & 0.2820 & 0.3244 & 0.3166 & \textbf{0.3586} & 0.3435 & 0.3362 & 0.2688 & 0.3344 \\
Colon & 0.3301 & 0.3482 & 0.3852 & 0.3527 & 0.3751 & 0.3817 & 0.3378 & \textbf{0.3932} \\
Esophagus & 0.3651 & 0.3860 & \textbf{0.4017} & 0.3768 & 0.3989 & 0.3873 & 0.3539 & 0.3977 \\
H\&N* & 0.3113 & 0.3251 & 0.3817 & 0.3430 & 0.3725 & \textbf{0.3884} & 0.3622 & 0.3548 \\
Kidney & 0.2473 & 0.2417 & 0.2506 & 0.2356 & \textbf{0.2726} & 0.2609 & 0.2045 & 0.2335 \\
Liver & 0.3698 & 0.3996 & 0.4068 & 0.4112 & 0.4003 & 0.4098 & 0.3695 & \textbf{0.4242} \\
Lung & 0.2480 & 0.2826 & 0.2834 & 0.2719 & 0.2700 & \textbf{0.2905} & 0.2547 & 0.2886 \\
Ovarian & 0.3509 & 0.3849 & \textbf{0.4296} & 0.4185 & 0.4037 & 0.4258 & 0.3852 & 0.4278 \\
Pancreatic & 0.2379 & 0.2952 & 0.3442 & 0.3020 & \textbf{0.3492} & 0.3332 & 0.2469 & 0.2530 \\
Prostate & 0.2465 & 0.3187 & \textbf{0.3260} & 0.2950 & 0.3415 & 0.3089 & 0.2526 & 0.3039 \\
Skin & 0.2093 & 0.2787 & 0.2718 & 0.2920 & 0.2780 & 0.3060 & 0.2624 & \textbf{0.3070} \\
Stomach & 0.3211 & 0.3232 & \textbf{0.3659} & 0.3323 & 0.3567 & 0.3725 & 0.3427 & 0.3506 \\
Testis & 0.3306 & 0.3544 & 0.3894 & 0.3876 & 0.3864 & 0.3752 & 0.3347 & \textbf{0.4051} \\
Thyroid & 0.2989 & 0.3250 & 0.3397 & 0.3007 & \textbf{0.3474} & 0.3441 & 0.3111 & 0.3312 \\
Uterus & 0.2354 & 0.2420 & 0.2575 & 0.2512 & 0.2548 & 0.2590 & 0.2385 & \textbf{0.2578} \\
\hline
Overall & \begin{tabular}[c]{@{}c@{}}0.3915\\ ±0.0028\end{tabular} & \begin{tabular}[c]{@{}c@{}}0.4147\\ ±0.0022\end{tabular} & \begin{tabular}[c]{@{}c@{}}0.4379\\ ±0.0043\end{tabular} & \begin{tabular}[c]{@{}c@{}}0.4230\\ ±0.0029\end{tabular} & \begin{tabular}[c]{@{}c@{}}0.4369\\ ±0.0021\end{tabular} & \begin{tabular}[c]{@{}c@{}}0.4375\\ ±0.0057\end{tabular} & \begin{tabular}[c]{@{}c@{}}0.3964\\ ±0.0038\end{tabular} & \begin{tabular}[c]{@{}c@{}}\textbf{0.4432}\\ ±0.0022\end{tabular} \\
\hline
\end{tabular}
}
\end{table}

\begin{table*}[ht]
\caption{\label{tab:semi_experiment}
Performance comparison among different self-supervised pretraining methods with the semi-supervised learning protocol. The number of images associated with each annotation budget is indicated in the parentheses.}
\resizebox{\textwidth}{!}{%
\begin{tabular}{lcccccccccccccccccccc}
\hline
\multicolumn{1}{c}{\multirow{3}{*}{Method}} & \multicolumn{6}{c}{Lizard-Dpath} &  & \multicolumn{6}{c}{Lizard-CRAG} &  & \multicolumn{6}{c}{Kather} \\
\multicolumn{1}{c}{} & \multicolumn{2}{c}{10\% (2)} & \multicolumn{2}{c}{25\% (5)} & \multicolumn{2}{c}{50\% (10)} &  & \multicolumn{2}{c}{10\% (2)} & \multicolumn{2}{c}{25\% (5)} & \multicolumn{2}{c}{50\% (10)} &  & \multicolumn{2}{c}{0.25\% (50)} & \multicolumn{2}{c}{0.50\% (88)} & \multicolumn{2}{c}{1.00\% (178)} \\ \cline{2-7} \cline{9-14} \cline{16-21} 
\multicolumn{1}{c}{} & mPQ+ & F1 & mPQ+ & F1 & mPQ+ & F1 &  & mPQ+ & F1 & mPQ+ & F1 & mPQ+ & F1 &  & Acc & F1 & Acc & F1 & Acc & F1 \\ \hline
\multicolumn{21}{l}{\textbf{Baseline}} \\
Random & \begin{tabular}[c]{@{}c@{}}0.1573\\ ±0.0039\end{tabular} & \begin{tabular}[c]{@{}c@{}}0.2119\\ ±0.073\end{tabular} & \begin{tabular}[c]{@{}c@{}}0.2667\\ ±0.0042\end{tabular} & \begin{tabular}[c]{@{}c@{}}0.3305\\ ±0.0059\end{tabular} & \begin{tabular}[c]{@{}c@{}}0.2901\\ ±0.0029\end{tabular} & \begin{tabular}[c]{@{}c@{}}0.3556\\ ±0.0067\end{tabular} &  & \begin{tabular}[c]{@{}c@{}}0.1912\\ ±0.0021\end{tabular} & \begin{tabular}[c]{@{}c@{}}0.2488\\ ±0.0036\end{tabular} & \begin{tabular}[c]{@{}c@{}}0.2311\\ ±0.0047\end{tabular} & \begin{tabular}[c]{@{}c@{}}0.2910\\ ±0.0059\end{tabular} & \begin{tabular}[c]{@{}c@{}}0.2709\\ ±0.0033\end{tabular} & \begin{tabular}[c]{@{}c@{}}0.3371\\ ±0.0041\end{tabular} &  & \begin{tabular}[c]{@{}c@{}}0.8021\\ ±0.0128\end{tabular} & \begin{tabular}[c]{@{}c@{}}0.8009\\ ±0.0117\end{tabular} & \begin{tabular}[c]{@{}c@{}}0.8399\\ ±0.0104\end{tabular} & \begin{tabular}[c]{@{}c@{}}0.8360\\ ±0.0101\end{tabular} & \begin{tabular}[c]{@{}c@{}}0.8620\\ ±0.0089\end{tabular} & \begin{tabular}[c]{@{}c@{}}0.8578\\ ±0.0091\end{tabular} \\ \hline
\multicolumn{21}{l}{\textbf{Generalized Pretraining Approaches}} \\
MoCo v2\cite{chen2020mocov2} & \begin{tabular}[c]{@{}c@{}}0.1569\\ ±0.0027\end{tabular} & \begin{tabular}[c]{@{}c@{}}0.2080\\ ±0.0044\end{tabular} & \begin{tabular}[c]{@{}c@{}}0.2679\\ ±0.0051\end{tabular} & \begin{tabular}[c]{@{}c@{}}0.3353\\ ±0.0063\end{tabular} & \begin{tabular}[c]{@{}c@{}}0.2898\\ ±0.0042\end{tabular} & \begin{tabular}[c]{@{}c@{}}0.3559\\ ±0.0060\end{tabular} &  & \begin{tabular}[c]{@{}c@{}}0.1940\\ ±0.0019\end{tabular} & \begin{tabular}[c]{@{}c@{}}0.2542\\ ±0.0028\end{tabular} & \begin{tabular}[c]{@{}c@{}}0.2410\\ ±0.0040\end{tabular} & \begin{tabular}[c]{@{}c@{}}0.3042\\ ±0.0052\end{tabular} & \begin{tabular}[c]{@{}c@{}}0.2795\\ ±0.0025\end{tabular} & \begin{tabular}[c]{@{}c@{}}0.3482\\ ±0.0032\end{tabular} &  & \begin{tabular}[c]{@{}c@{}}0.8466\\ ±0.0121\end{tabular} & \begin{tabular}[c]{@{}c@{}}0.8439\\ ±0.0118\end{tabular} & \begin{tabular}[c]{@{}c@{}}0.8881\\ ±0.0110\end{tabular} & \begin{tabular}[c]{@{}c@{}}0.8869\\ ±0.0109\end{tabular} & \begin{tabular}[c]{@{}c@{}}0.9296\\ ±0.0052\end{tabular} & \begin{tabular}[c]{@{}c@{}}0.9326\\ ±0.0062\end{tabular} \\
Simsiam\cite{Chen2021SimSia} & \begin{tabular}[c]{@{}c@{}}0.1557\\ ±0.0037\end{tabular} & \begin{tabular}[c]{@{}c@{}}0.2036\\ ±0.0058\end{tabular} & \begin{tabular}[c]{@{}c@{}}0.2630\\ ±0.0033\end{tabular} & \begin{tabular}[c]{@{}c@{}}0.3283\\ ±0.0041\end{tabular} & \begin{tabular}[c]{@{}c@{}}0.2920\\ ±0.0021\end{tabular} & \begin{tabular}[c]{@{}c@{}}0.3616\\ ±0.0037\end{tabular} &  & \begin{tabular}[c]{@{}c@{}}0.1911\\ ±0.0015\end{tabular} & \begin{tabular}[c]{@{}c@{}}0.2476\\ ±0.0026\end{tabular} & \begin{tabular}[c]{@{}c@{}}0.2439\\ ±0.0051\end{tabular} & \begin{tabular}[c]{@{}c@{}}0.3044\\ ±0.0066\end{tabular} & \begin{tabular}[c]{@{}c@{}}0.2761\\ ±0.0022\end{tabular} & \begin{tabular}[c]{@{}c@{}}0.3417\\ ±0.0029\end{tabular} &  & \begin{tabular}[c]{@{}c@{}}0.8021\\ ±0.0069\end{tabular} & \begin{tabular}[c]{@{}c@{}}0.8007\\ ±0.0076\end{tabular} & \begin{tabular}[c]{@{}c@{}}0.8365\\ ±0.0055\end{tabular} & \begin{tabular}[c]{@{}c@{}}0.8339\\ ±0.0049\end{tabular} & \begin{tabular}[c]{@{}c@{}}0.8910\\ ±0.0049\end{tabular} & \begin{tabular}[c]{@{}c@{}}0.8776\\ ±0.0046\end{tabular} \\ \hline
\multicolumn{21}{l}{\textbf{Pretraining Approaches for Medical Images}} \\
DiRA\cite{Haghighi2022DiRA} & \begin{tabular}[c]{@{}c@{}}0.1421\\ ±0.0021\end{tabular} & \begin{tabular}[c]{@{}c@{}}0.1903\\ ±0.0039\end{tabular} & \begin{tabular}[c]{@{}c@{}}0.2547\\ ±0.0029\end{tabular} & \begin{tabular}[c]{@{}c@{}}0.3184\\ ±0.0039\end{tabular} & \begin{tabular}[c]{@{}c@{}}0.2930\\ ±0.0033\end{tabular} & \begin{tabular}[c]{@{}c@{}}0.36231\\ ±0.0045\end{tabular} &  & \begin{tabular}[c]{@{}c@{}}0.1909\\ ±0.0020\end{tabular} & \begin{tabular}[c]{@{}c@{}}0.2523\\ ±0.0029\end{tabular} & \begin{tabular}[c]{@{}c@{}}0.2334\\ ±0.0046\end{tabular} & \begin{tabular}[c]{@{}c@{}}0.2962\\ ±0.0061\end{tabular} & \begin{tabular}[c]{@{}c@{}}0.2752\\ ±0.0029\end{tabular} & \begin{tabular}[c]{@{}c@{}}0.3428\\ ±0.0038\end{tabular} &  & \begin{tabular}[c]{@{}c@{}}\textbf{0.9001}\\ ±0.0112\end{tabular} & \begin{tabular}[c]{@{}c@{}}\textbf{0.8980}\\ ±0.0109\end{tabular} & \begin{tabular}[c]{@{}c@{}}\underline{0.9230}\\ ±0.0106\end{tabular} & \begin{tabular}[c]{@{}c@{}}\underline{0.9205}\\ ±0.0107\end{tabular} & \begin{tabular}[c]{@{}c@{}}\underline{0.9530}\\ ±0.0069\end{tabular} & \begin{tabular}[c]{@{}c@{}}\underline{0.9505}\\ ±0.0081\end{tabular} \\
Ciga\cite{ciga2022contrastivepath} & \begin{tabular}[c]{@{}c@{}}\underline{0.1688}\\ ±0.0042\end{tabular} & \begin{tabular}[c]{@{}c@{}}\underline{0.2180}\\ ±0.0065\end{tabular} & \begin{tabular}[c]{@{}c@{}}\underline{0.2859}\\ ±0.0044\end{tabular} & \begin{tabular}[c]{@{}c@{}}\underline{0.3521}\\ ±0.0067\end{tabular} & \begin{tabular}[c]{@{}c@{}}\underline{0.3118}\\ ±0.0039\end{tabular} & \begin{tabular}[c]{@{}c@{}}\underline{0.3839}\\ ±0.0049\end{tabular} &  & \begin{tabular}[c]{@{}c@{}}\underline{0.2120}\\ ±0.0012\end{tabular} & \begin{tabular}[c]{@{}c@{}}\underline{0.2746}\\ ±0.0022\end{tabular} & \begin{tabular}[c]{@{}c@{}}\underline{0.2522}\\ ±0.0038\end{tabular} & \begin{tabular}[c]{@{}c@{}}\underline{0.3166}\\ ±0.0050\end{tabular} & \begin{tabular}[c]{@{}c@{}}\underline{0.2844}\\ ±0.0024\end{tabular} & \begin{tabular}[c]{@{}c@{}}\underline{0.3510}\\ ±0.0030\end{tabular} &  & \begin{tabular}[c]{@{}c@{}}0.8201\\ ±0.0081\end{tabular} & \begin{tabular}[c]{@{}c@{}}0.8195\\ ±0.0099\end{tabular} & \begin{tabular}[c]{@{}c@{}}0.8497\\ ±0.072\end{tabular} & \begin{tabular}[c]{@{}c@{}}0.8447\\ ±0.079\end{tabular} & \begin{tabular}[c]{@{}c@{}}0.8699\\ ±0.0088\end{tabular} & \begin{tabular}[c]{@{}c@{}}0.8642\\ ±0.0081\end{tabular} \\
\rowcolor{gray} Ours & \begin{tabular}[c]{@{}c@{}}\textbf{0.2138}\\ ±0.0011\end{tabular} & \begin{tabular}[c]{@{}c@{}}\textbf{0.2697}\\ ±0.0023\end{tabular} & \begin{tabular}[c]{@{}c@{}}\textbf{0.3211}\\ ±0.0049\end{tabular} & \begin{tabular}[c]{@{}c@{}}\textbf{0.4015}\\ ±0.0062\end{tabular} & \begin{tabular}[c]{@{}c@{}}\textbf{0.3411}\\ ±0.0013\end{tabular} & \begin{tabular}[c]{@{}c@{}}\textbf{0.4216}\\ ±0.0023\end{tabular} &  & \begin{tabular}[c]{@{}c@{}}\textbf{0.2478}\\ ±0.0017\end{tabular} & \begin{tabular}[c]{@{}c@{}}\textbf{0.3119}\\ ±0.0025\end{tabular} & \begin{tabular}[c]{@{}c@{}}\textbf{0.2710}\\ ±0.0035\end{tabular} & \begin{tabular}[c]{@{}c@{}}\textbf{0.3364}\\ ±0.0065\end{tabular} & \begin{tabular}[c]{@{}c@{}}\textbf{0.3217}\\ ±0.0020\end{tabular} & \begin{tabular}[c]{@{}c@{}}\textbf{0.3938}\\ ±0.0029\end{tabular} &  & \begin{tabular}[c]{@{}c@{}}\underline{0.8967}\\ ±0.0095\end{tabular} & \begin{tabular}[c]{@{}c@{}}\underline{0.8934}\\ ±0.0101\end{tabular} & \begin{tabular}[c]{@{}c@{}}\textbf{0.9339}\\ ±0.0033\end{tabular} & \begin{tabular}[c]{@{}c@{}}\textbf{0.9328}\\ ±0.0062\end{tabular} & \begin{tabular}[c]{@{}c@{}}\textbf{0.9539}\\ ±0.0041\end{tabular} & \begin{tabular}[c]{@{}c@{}}\textbf{0.9508}\\ ±0.0040\end{tabular} \\ \hline
\end{tabular}%
}
\end{table*}

\begin{figure*}[t]
	\centering
	\includegraphics[width=\textwidth]{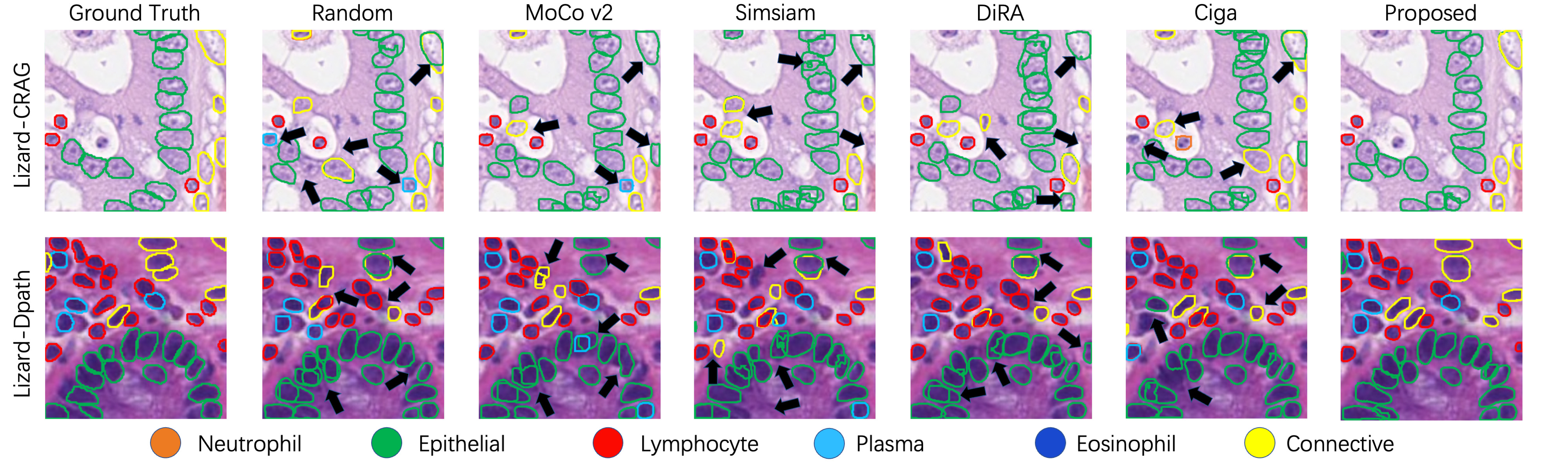}
\caption{Qualitative results of different pretraining approaches for semi-supervised nuclear segmentation and detection with the annotation budget of 25\%. The boundary colour of each nuclear denotes its category.}
	\label{comparison_semi}
\end{figure*}

\subsubsection{Colorectal Cancer Classification}
Cancer classification is performed on the Kather dataset.
We only fine-tune on 1\% training data to make the task more challenging and exacerbate the differences amongst pretraining approaches.
Results in \autoref{tab:transfer_experiment} indicate that ImageNet provides more effective initialization than most of the pretraining methods.
The results are in line with the finding that the self-supervised methods do not always achieve superior results against ImageNet \cite{ciga2022contrastivepath}.
Despite this, 
our method not only significantly outperforms other self-supervised methods with an increase of at least 1.12\% for accuracy and 1.08\% for F1 score (0.9590 \emph{vs.} 0.9478 and 0.9564 \emph{vs.} 0.9456, compared with MoCo v2, p$<$0.05), but also achieves slightly better results than ImageNet.
One possible explanation for these results could be that the cancer classification task is highly dependent on the properties of nuclei, which our pretraining method can help recognize better.

\subsubsection{Gland Segmentation}
\label{gland_seg}
We perform gland segmentation tasks on the Crag dataset.
It is noteworthy that although our pretraining method is nucleus-aware, we also provide stronger initialization for gland segmentation than supervised counterparts (ImageNet, p$<$0.05 and Mormont, p$<$0.05, Obj Dice).
We also achieve superior results than other self-supervised methods, although the improvement is not significant (0.8649 \emph{vs.} 0.8616, p=0.28, compared with Ciga).
The improvements can be explained by our special design of the nuclear distribution in our mask images, which is similar to that of epithelial cells around the glands.

\subsubsection{Multi-class Nuclear Detection and Segmentation}
On the Dpath subset of Lizard, the proposed method outperforms other pretraining methods for mPQ+.
However, no improvement in F1 score is observed compared to MoCo v2.
We attribute the results to the trade-off between nuclear detection and subtyping, which is more obvious in the proposed approach because we do not classify the instances in our segmentation branch during pretraining.
When experimenting on the CRAG subset of Lizard, we achieve the best results in terms of both mPQ+ and F1.

\subsubsection{Multi-organ and Multi-class Nuclear Detection and Segmentation}
We also experiment on a more diverse dataset to further evaluate the transferability of our pretrained model in case multi-organ and multi-class nuclei need to be segmented.
The experiments performed on PanNuke indicate that our method outperforms DiRA (p$<$0.05), which is the previous SOTA pretraining method.
We can learn from the per-organ results in \autoref{tab:pannuke_results} that both the epithelial-like structures (\emph{e.g.}, those in breast, colon, and liver) and non-epithelial-like structures (\emph{e.g.}, those in bladder) benefit from our pretraining methods.
The results suggest the robustness of our pretrained model that benefits various downstream situations.

\subsubsection{Multiple Instance Learning}
The multiple instance learning experiment is performed on the In-house-MIL dataset to evaluate whether our method benefits the special two-stage model for MIL.
The original work \cite{sharma2021c2c} used ImageNet-pretrained backbone for sampling and aggregation.
We also find the strategy effective compared with initializing the backbone randomly, which can hardly provide discriminative features after sampling.
Moreover, it is observed that the generalized self-supervised pretraining methods fail to provide stronger initialization, whereas the pertaining methods for medical images benefit more compared with ImageNet.
In such a special task, our method still demonstrates its superiority compared with ImageNet (p$<$0.05), and outperforms other pretraining approaches.

\subsection{Semi-supervised Learning Experiments}
Let $D_l = \{( \mathbf{x}_i, y_i)\}_{i=1}^{N}$ and $D_u = \{( \mathbf{x}_i)\}_{i=1}^{M}$ denote the labeled samples and unlabeled samples from the same distribution ($M \gg N$).
Semi-supervised learning (SSL) aims to utilize $D_u$ to boost the model performance on $D_l$.
We conduct SSL following the protocol that self-supervised pretraining is performed on the whole training set and fine-tuning is later performed on parts of them\cite{Koohbanani2021SelfPath}.
We compare with other self-supervised pretraining approaches described in \ref{transfer_experiment} with the same protocol.
To evaluate the performance, we vary the annotation budget for the training set, keep the validation set fixed and report the performance on the test set.

\begin{figure}[t]
	\centering
	\includegraphics[width=\columnwidth]{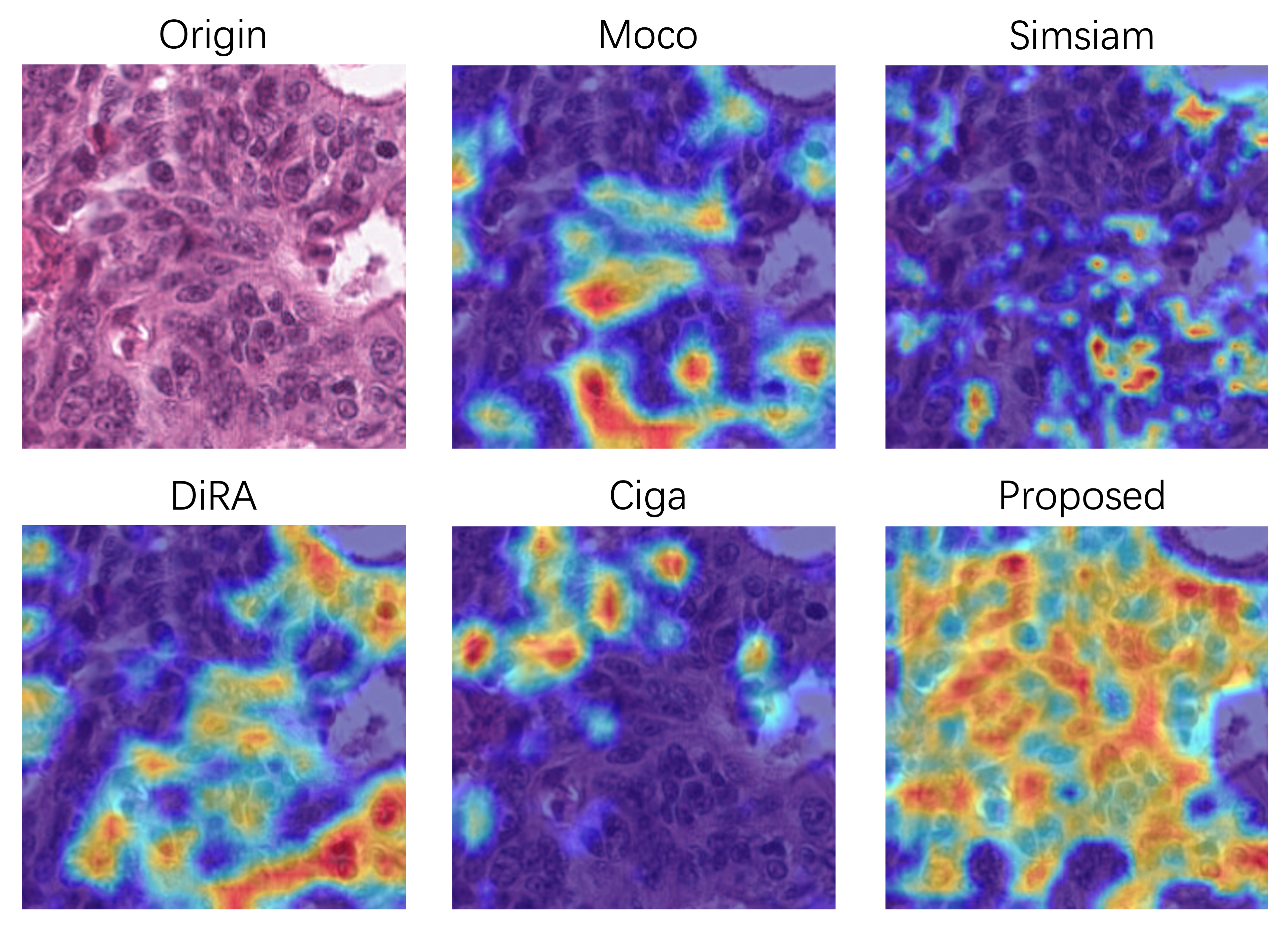}
\caption{Grad-Cam results for semi-supervised classification with 0.5\% annotation budget.}
	\label{grad_semi}
\end{figure}

\subsubsection{Multi-class Nuclear Detection and Segmentation}
We use the CRAG subset and the Dpath subset of Lizard for the nuclear object detection and segmentation experiment.
We fine-tune Mask-RCNN with different annotation budgets (10\%, 25\%, and 50\%) and present the results for each subset in \autoref{tab:semi_experiment}.
Our pretraining approach improves the performance on both subsets and all of the metrics over prior methods significantly (p$<$0.05).
On the CRAG subset, it is striking to find that we just need 5\% of labeled data to match the best performance obtained from other methods that need 25\% annotation budgets (0.2478 \emph{vs.} 0.2522, mPQ+). 
Our method also performs well on the Dpath subset, where MoCo v2, Simsiam, and DiRA fail to provide effective pretrained models in most cases, whose results are even worse than random initialization.
We attribute the decreased performance of these approaches to insufficient pretraining data.
Specifically, these approaches may be overfitted to the pretext tasks in the Lizard training dataset, which contains only 194 images for pretraining.
The proposed pretraining method, on the other hand, does not have this issue and yields more robust results.

\subsubsection{Cancer Classification}
We pretrain on the whole training set and validation set of Kather without any annotations, and fine-tune the classifier with different annotation budgets (0.2\%, 0.5\%, and 1\%).
Similar to \cite{Koohbanani2021SelfPath}, all of the data regimes are extremely low due to the high baseline performance (0.9532 for Acc with 2\% training data).
\autoref{tab:semi_experiment} demonstrates that we outperform the baseline whose weights are initialized randomly (p$<$0.05).
It is noteworthy that our pretraining method focuses on fine-grained details where most of the nuclei are recognized.
This is confirmed in \autoref{grad_semi} with Grad-Cam \cite{GradCam} visualization algorithm which generates a heatmap that highlights supportive pixels in the histopathology image for the classifier.
It is also interesting to find that we not only focus on nucleus-level features, but also capture cell/tissue-level features, which may interpret the superior performance on gland segmentation in \ref{gland_seg}.
Such multi-level features are beneficial for classification, where we outperform other discriminative self-supervised learning approaches (\emph{i.e.,} MoCo, Simsiam, Ciga) significantly (p$<$0.05).
However, the performance boost is not as significant as segmentation tasks compared with DiRA, which captures stronger global representation by incorporating discriminative self-supervised learning with generative and adversarial ones.

\subsection{Ablation Study}
A thorough ablation study is conducted to justify the benefit of our design by progressively applying each component for our transfer learning and semi-supervised learning protocols (25\% training data).
We modify the method\cite{Mahmood2019GANseg2} as the baseline, which also uses a CycleGAN-based framework to translate between mask images and histopathology images.
It is noteworthy that their intention is to augment the training data for nuclear segmentation, which cannot be directly used for comparison.
Therefore, we replace their mask generator with our ResNet-based generator and only implement their CycleGAN stage for a fair comparison.

\begin{table}[t]
\centering
\caption{\label{tab:ablate_architecture}
Main results for ablation study. Semi-supervised learning is performed with 25 \% annotation budget.}
\resizebox{\columnwidth}{!}{%
\begin{tabular}{lccccccccc}
\hline
\multicolumn{1}{c}{\multirow{3}{*}{Configuration}} & \multicolumn{4}{c}{Transfer Learning} &  & \multicolumn{4}{c}{Semi-supervised Learning} \\
\multicolumn{1}{c}{} & \multicolumn{2}{c}{Kumar} & \multicolumn{2}{c}{Kather} &  & \multicolumn{2}{c}{Lizard-Dpath} & \multicolumn{2}{c}{Lizard-CRAG} \\ \cline{2-5} \cline{7-10} 
\multicolumn{1}{c}{} & AJI & F1 & Acc & F1 &  & mPQ+ & F1 & mPQ+ & F1 \\ \hline
\multicolumn{10}{l}{\textbf{Mask Images}} \\
Baseline\cite{Mahmood2019GANseg2} & \begin{tabular}[c]{@{}c@{}}0.5415\\ ±0.0039\end{tabular} & \begin{tabular}[c]{@{}c@{}}0.6805\\ ±0.0073\end{tabular} & \begin{tabular}[c]{@{}c@{}}0.8102\\ ±0.0063\end{tabular} & \begin{tabular}[c]{@{}c@{}}0.8159\\ ±0.0075\end{tabular} &  & \begin{tabular}[c]{@{}c@{}}0.2877\\ ±0.0034\end{tabular} & \begin{tabular}[c]{@{}c@{}}0.3587\\ ±0.0042\end{tabular} & \begin{tabular}[c]{@{}c@{}}0.2397\\ ±0.0053\end{tabular} & \begin{tabular}[c]{@{}c@{}}0.2982\\ ±0.0072\end{tabular} \\
+Distribution & \begin{tabular}[c]{@{}c@{}}0.5540\\ ±0.0024\end{tabular} & \begin{tabular}[c]{@{}c@{}}0.7020\\ ±0.0043\end{tabular} & \begin{tabular}[c]{@{}c@{}}0.8392\\ ±0.0052\end{tabular} & \begin{tabular}[c]{@{}c@{}}0.8359\\ ±0.0066\end{tabular} &  & \begin{tabular}[c]{@{}c@{}}0.2944\\ ±0.0058\end{tabular} & \begin{tabular}[c]{@{}c@{}}0.3705\\ ±0.0061\end{tabular} & \begin{tabular}[c]{@{}c@{}}0.2537\\ ±0.0035\end{tabular} & \begin{tabular}[c]{@{}c@{}}0.3185\\ ±0.0049\end{tabular} \\
+Stylization & \begin{tabular}[c]{@{}c@{}}0.5527\\ ±0.0044\end{tabular} & \begin{tabular}[c]{@{}c@{}}0.7121\\ ±0.0056\end{tabular} & \begin{tabular}[c]{@{}c@{}}0.8410\\ ±0.0026\end{tabular} & \begin{tabular}[c]{@{}c@{}}0.8396\\ ±0.0039\end{tabular} &  & \begin{tabular}[c]{@{}c@{}}0.3008\\ ±0.0031\end{tabular} & \begin{tabular}[c]{@{}c@{}}0.3785\\ ±0.0039\end{tabular} & \begin{tabular}[c]{@{}c@{}}0.2603\\ ±0.0055\end{tabular} & \begin{tabular}[c]{@{}c@{}}0.3233\\ ±0.0070\end{tabular} \\ \hline
\multicolumn{10}{l}{\textbf{Architecture and Training Strategy}} \\
+Two-stage & \begin{tabular}[c]{@{}c@{}}0.5614\\ ±0.0038\end{tabular} & \begin{tabular}[c]{@{}c@{}}0.7212\\ ±0.0058\end{tabular} & \begin{tabular}[c]{@{}c@{}}0.9227\\ ±0.0045\end{tabular} & \begin{tabular}[c]{@{}c@{}}0.9216\\ ±0.0070\end{tabular} &  & \begin{tabular}[c]{@{}c@{}}0.3071\\ ±0.0046\end{tabular} & \begin{tabular}[c]{@{}c@{}}0.3844\\ ±0.0055\end{tabular} & \begin{tabular}[c]{@{}c@{}}0.2619\\ ±0.0023\end{tabular} & \begin{tabular}[c]{@{}c@{}}0.3251\\ ±0.0029\end{tabular} \\
+Co-modulation & \begin{tabular}[c]{@{}c@{}}0.5801\\ ±0.0046\end{tabular} & \begin{tabular}[c]{@{}c@{}}0.7381\\ ±0.0059\end{tabular} & \begin{tabular}[c]{@{}c@{}}0.9419\\ ±0.0019\end{tabular} & \begin{tabular}[c]{@{}c@{}}0.9412\\ ±0.0023\end{tabular} &  & \begin{tabular}[c]{@{}c@{}}0.3011\\ ±0.0036\end{tabular} & \begin{tabular}[c]{@{}c@{}}0.3799\\ ±0.0047\end{tabular} & \begin{tabular}[c]{@{}c@{}}0.2526\\ ±0.0031\end{tabular} & \begin{tabular}[c]{@{}c@{}}0.3161\\ ±0.0047\end{tabular} \\
+ADA & \begin{tabular}[c]{@{}c@{}}0.5854\\ ±0.0019\end{tabular} & \begin{tabular}[c]{@{}c@{}}0.7423\\ ±0.0037\end{tabular} & \begin{tabular}[c]{@{}c@{}}0.9492\\ ±0.0020\end{tabular} & \begin{tabular}[c]{@{}c@{}}0.9429\\ ±0.0026\end{tabular} &  & \begin{tabular}[c]{@{}c@{}}0.3130\\ ±0.0030\end{tabular} & \begin{tabular}[c]{@{}c@{}}0.3912\\ ±0.0056\end{tabular} & \begin{tabular}[c]{@{}c@{}}0.2701\\ ±0.0044\end{tabular} & \begin{tabular}[c]{@{}c@{}}0.3354\\ ±0.0062\end{tabular} \\
\rowcolor{gray} +ISG & \begin{tabular}[c]{@{}c@{}}\textbf{0.5932}\\ ±0.0025\end{tabular} & \begin{tabular}[c]{@{}c@{}}\textbf{0.7591}\\ ±0.0044\end{tabular} & \begin{tabular}[c]{@{}c@{}}\textbf{0.9588}\\ ±0.0028\end{tabular} & \begin{tabular}[c]{@{}c@{}}\textbf{0.9560}\\ ±0.0041\end{tabular} &  & \begin{tabular}[c]{@{}c@{}}\textbf{0.3211}\\ ±0.0049\end{tabular} & \begin{tabular}[c]{@{}c@{}}\textbf{0.4015}\\ ±0.0062\end{tabular} & \begin{tabular}[c]{@{}c@{}}\textbf{0.2710}\\ ±0.0035\end{tabular} & \begin{tabular}[c]{@{}c@{}}\textbf{0.3364}\\ ±0.0065\end{tabular} \\ \hline
\end{tabular}%
}
\end{table}

\begin{table}[t]
\caption{\label{tab:ablate_sg}
Ablation study on the segmentation guided strategy.}
\resizebox{\columnwidth}{!}{%
\begin{tabular}{lccccccccc}
\hline
\multicolumn{1}{c}{\multirow{3}{*}{Configuration}} & \multicolumn{4}{c}{Transfer Learning} &  & \multicolumn{4}{c}{Semi-supervised Learning} \\
\multicolumn{1}{c}{} & \multicolumn{2}{c}{Kumar} & \multicolumn{2}{c}{Kather} &  & \multicolumn{2}{c}{Lizard-Dpath} & \multicolumn{2}{c}{Lizard-CRAG} \\ \cline{2-5} \cline{7-10} 
\multicolumn{1}{c}{} & AJI & F1 & Acc & F1 &  & mPQ+ & F1 & mPQ+ & F1 \\ \hline
w/o SG & \begin{tabular}[c]{@{}c@{}}0.5854\\ ±0.0019\end{tabular} & \begin{tabular}[c]{@{}c@{}}0.7423\\ ±0.0037\end{tabular} & \begin{tabular}[c]{@{}c@{}}0.9492\\ ±0.0020\end{tabular} & \begin{tabular}[c]{@{}c@{}}0.9429\\ ±0.0026\end{tabular} &  & \begin{tabular}[c]{@{}c@{}}0.3130\\ ±0.0030\end{tabular} & \begin{tabular}[c]{@{}c@{}}0.3912\\ ±0.0056\end{tabular} & \begin{tabular}[c]{@{}c@{}}0.2701\\ ±0.0044\end{tabular} & \begin{tabular}[c]{@{}c@{}}0.3354\\ ±0.0062\end{tabular} \\
w/ SSG & \begin{tabular}[c]{@{}c@{}}0.5768\\ ±0.0033\end{tabular} & \begin{tabular}[c]{@{}c@{}}0.7280\\ ±0.0047\end{tabular} & \begin{tabular}[c]{@{}c@{}}0.9477\\ ±0.0024\end{tabular} & \begin{tabular}[c]{@{}c@{}}0.9436\\ ±0.0045\end{tabular} &  & \begin{tabular}[c]{@{}c@{}}0.3101\\ ±0.0053\end{tabular} & \begin{tabular}[c]{@{}c@{}}0.3882\\ ±0.0072\end{tabular} & \begin{tabular}[c]{@{}c@{}}0.2539\\ ±0.0033\end{tabular} & \begin{tabular}[c]{@{}c@{}}0.3168\\ ±0.0059\end{tabular} \\ 
\rowcolor{gray} w/ ISG & \begin{tabular}[c]{@{}c@{}}\textbf{0.5932}\\ ±0.0025\end{tabular} & \begin{tabular}[c]{@{}c@{}}\textbf{0.7591}\\ ±0.0044\end{tabular} & \begin{tabular}[c]{@{}c@{}}\textbf{0.9588}\\ ±0.0028\end{tabular} & \begin{tabular}[c]{@{}c@{}}\textbf{0.9560}\\ ±0.0041\end{tabular} &  & \begin{tabular}[c]{@{}c@{}}\textbf{0.3211}\\ ±0.0049\end{tabular} & \begin{tabular}[c]{@{}c@{}}\textbf{0.4015}\\ ±0.0062\end{tabular} & \begin{tabular}[c]{@{}c@{}}\textbf{0.2710}\\ ±0.0035\end{tabular} & \begin{tabular}[c]{@{}c@{}}\textbf{0.3364}\\ ±0.0065\end{tabular} \\ \hline
\end{tabular}%
}
\end{table}

\subsubsection{Mask Image Quality}
We examine the effect of the quality of synthesized mask images.
As illustrated in the corresponding section in \autoref{tab:ablate_architecture}, extra information brought by the nuclei distribution and stylization is introduced.
We begin with a random distribution and binary nuclei mask design following \cite{Mahmood2019GANseg2}, then progressively add the glandular structure and stylize the nuclei.
Generally, we can observe the performance gain for both the transfer learning and semi-supervised learning with the introduction of distribution priors (p$<$0.05).
Stylization benefits most cases, but slightly degrades the segmentation performance on Kumar.

\subsubsection{Architecture design}
To analyze the effect of the network design, we investigate different architectures of our framework.
As summarized in \autoref{tab:ablate_architecture}, we divide the design into three main parts: The co-modulation design for pathology generator $G$, the adaptive discriminator augmentation strategy (ADA) for $D_G$, and the segmentation guided strategy.
For the first part, we find it effective in our transfer learning protocols.
However, in our semi-supervised learning protocols where the data for pretraining is limited, the complex generator design is not beneficial due to the over-fitting problem.
We alleviate the problem by augmenting the pretraining data with ADA, which effectively boosts the performance for semi-supervised learning, while also further improving the performance for transfer learning.
Moreover, we discuss the benefits of instance segmentation guided strategy (ISG).
The results in \autoref{tab:ablate_architecture} and \autoref{tab:ablate_sg} show that the semantic segmentation guided strategy fails to offer useful information, supporting our analysis in \ref{sub:sg}.
On the contrary, the instance segmentation guided strategy effectively improves the performance, especially for transfer learning experiments where the improvement of 1.68\% is observed on Kumar and the improvement of 1.31 \% is observed on Kather (both in terms of F1 score, p$<$0.05).

\subsubsection{Pretraining Schedule}
Due to the inconsistent convergence rates of $G$ and $S$, we design a two-stage pretraining strategy. 
\autoref{tab:ablate_architecture} and \autoref{tab:ablate_strategy} investigate the effect of the extra training stages.
Overall, the additional training stage yields superior performance than pretraining for a single stage (p$<$0.05 for transfer learning).
We observe no significant performance improvement with more training stages.

\begin{table}[t]
\caption{\label{tab:ablate_strategy}
Ablation study on the pretraining schedule.}
\resizebox{\columnwidth}{!}{%
\begin{tabular}{lccccccccc}
\hline
\multicolumn{1}{c}{\multirow{3}{*}{Configuration}} & \multicolumn{4}{c}{Transfer Learning} &  & \multicolumn{4}{c}{Semi-supervised Learning} \\
\multicolumn{1}{c}{} & \multicolumn{2}{c}{Kumar} & \multicolumn{2}{c}{Kather} &  & \multicolumn{2}{c}{Lizard-Dpath} & \multicolumn{2}{c}{Lizard-CRAG} \\ \cline{2-5} \cline{7-10} 
\multicolumn{1}{c}{} & AJI & F1 & Acc & F1 &  & mPQ+ & F1 & mPQ+ & F1 \\ \hline
One-stage & \begin{tabular}[c]{@{}c@{}}0.5844\\ ±0.0056\end{tabular} & \begin{tabular}[c]{@{}c@{}}0.7409\\ ±0.0088\end{tabular} & \begin{tabular}[c]{@{}c@{}}0.8661\\ ±0.0036\end{tabular} & \begin{tabular}[c]{@{}c@{}}0.8585\\ ±0.0040\end{tabular} &  & \begin{tabular}[c]{@{}c@{}}0.3156\\ ±0.0043\end{tabular} & \begin{tabular}[c]{@{}c@{}}0.3956\\ ±0.0055\end{tabular} & \begin{tabular}[c]{@{}c@{}}0.2692\\ ±0.0027\end{tabular} & \begin{tabular}[c]{@{}c@{}}0.3356\\ ±0.0049\end{tabular} \\
\rowcolor{gray} Two-stage & \begin{tabular}[c]{@{}c@{}}\textbf{0.5932}\\ ±0.0025\end{tabular} & \begin{tabular}[c]{@{}c@{}}\textbf{0.7591}\\ ±0.0044\end{tabular} & \begin{tabular}[c]{@{}c@{}}\textbf{0.9588}\\ ±0.0028\end{tabular} & \begin{tabular}[c]{@{}c@{}}\textbf{0.9560}\\ ±0.0041\end{tabular} &  & \begin{tabular}[c]{@{}c@{}}\textbf{0.3211}\\ ±0.0049\end{tabular} & \begin{tabular}[c]{@{}c@{}}\textbf{0.4015}\\ ±0.0062\end{tabular} & \begin{tabular}[c]{@{}c@{}}0.2710\\ ±0.0035\end{tabular} & \begin{tabular}[c]{@{}c@{}}\textbf{0.3364}\\ ±0.0065\end{tabular} \\
Three-stage & \begin{tabular}[c]{@{}c@{}}0.5928\\ ±0.0043\end{tabular} & \begin{tabular}[c]{@{}c@{}}0.7546\\ ±0.0055\end{tabular} & \begin{tabular}[c]{@{}c@{}}0.9280\\ ±0.0026\end{tabular} & \begin{tabular}[c]{@{}c@{}}0.9311\\ ±0.0037\end{tabular} &  & \begin{tabular}[c]{@{}c@{}}0.3187\\ ±0.0054\end{tabular} & \begin{tabular}[c]{@{}c@{}}0.3993\\ ±0.0062\end{tabular} & \begin{tabular}[c]{@{}c@{}}\textbf{0.2714}\\ ±0.0042\end{tabular} & \begin{tabular}[c]{@{}c@{}}0.3356\\ ±0.52\end{tabular} \\ \hline
\end{tabular}%
}
\end{table}

\begin{table}[t]
\centering
\caption{\label{tab:ablate_initialize}
Ablation study on the initializing strategy. We are able to initialize the FPN and the instance segmentation branch, which is indicated by "all available".}
\resizebox{\columnwidth}{!}{%
\begin{tabular}{lccccccccc}
\hline
\multicolumn{1}{c}{\multirow{3}{*}{Initialized}} & \multicolumn{4}{c}{Transfer Learning} &  & \multicolumn{4}{c}{Semi-supervised Learning} \\
\multicolumn{1}{c}{} & \multicolumn{2}{c}{Kumar} & \multicolumn{2}{c}{Kather} &  & \multicolumn{2}{c}{Lizard-Dpath} & \multicolumn{2}{c}{Lizard-CRAG} \\ \cline{2-5} \cline{7-10} 
\multicolumn{1}{c}{} & AJI & F1 & Acc & F1 &  & mPQ+ & F1 & mPQ+ & F1 \\ \hline
ResNet & \begin{tabular}[c]{@{}c@{}}0.5881\\ ±0.0042\end{tabular} & \begin{tabular}[c]{@{}c@{}}0.7579\\ ±0.0066\end{tabular} & \begin{tabular}[c]{@{}c@{}}\textbf{0.9588}\\ ±0.0028\end{tabular} & \begin{tabular}[c]{@{}c@{}}\textbf{0.9560}\\ ±0.0041\end{tabular} &  & \begin{tabular}[c]{@{}c@{}}0.3085\\ ±0.0043\end{tabular} & \begin{tabular}[c]{@{}c@{}}0.3931\\ ±0.0052\end{tabular} & \begin{tabular}[c]{@{}c@{}}0.2696\\ ±0.0029\end{tabular} & \begin{tabular}[c]{@{}c@{}}0.3290\\ ±0.0058\end{tabular} \\
\rowcolor{gray} FPN & \begin{tabular}[c]{@{}c@{}}0.5932\\ ±0.0025\end{tabular} & \begin{tabular}[c]{@{}c@{}}0.7591\\ ±0.0044\end{tabular} & - & - &  & \begin{tabular}[c]{@{}c@{}}\textbf{0.3211}\\ ±0.0049\end{tabular} & \begin{tabular}[c]{@{}c@{}}\textbf{0.4015}\\ ±0.0062\end{tabular} & \begin{tabular}[c]{@{}c@{}}\textbf{0.2710}\\ ±0.0035\end{tabular} & \begin{tabular}[c]{@{}c@{}}\textbf{0.3364}\\ ±0.0065\end{tabular} \\
All available & \begin{tabular}[c]{@{}c@{}}\textbf{0.5966}\\ ±0.0031\end{tabular} & \begin{tabular}[c]{@{}c@{}}\textbf{0.7620}\\ ±0.0048\end{tabular}  & - & - &  & \begin{tabular}[c]{@{}c@{}}0.3019\\ ±0.0037\end{tabular} & \begin{tabular}[c]{@{}c@{}}0.3829\\ ±0.0049\end{tabular} & \begin{tabular}[c]{@{}c@{}}0.2533\\ ±0.0031\end{tabular} & \begin{tabular}[c]{@{}c@{}}0.3192\\ ±0.0047\end{tabular} \\ \hline
\end{tabular}%
}
\end{table}

\subsubsection{Network Initialization for Dense-prediction Tasks}
We evaluate the effect of initializing the ResNet backbone, FPN, and all the structures we can initialize (including the FPN and the instance segmentation branch) in \autoref{tab:ablate_initialize}.
Initializing FPN demonstrates better performance than merely initializing the encoder backbone.
It is also inspiring to find that the performance can be further improved by further initializing the instance segmentation head on Kumar.
However, when additional instance segmentation heads are further initialized, we observe a minor improvement on Kather but degraded performance on Lizard.
This may be attributed to the different roles of the segmentation branch during pretraining and fine-tuning.
In downstream tasks, when the segmentation branch further classifies the nuclei, initializing the heads of the segmenter might cause the network to ignore the differences among nuclei.
Therefore, we only initialize the FPN for all the dense-prediction tasks.

\section{Conclusion and Discussion}
In this paper, a nucleus-aware self-supervised framework based on UNIT is introduced for histopathology images.
Due to the importance of nuclear distribution and morphology for pathologic analysis,
it is a requirement that a self-supervised learning framework includes these priors.
In our method, the cycle consistency between histopathology images and pseudo mask images containing rich information about the nuclei is intended to impose the model's awareness of nuclear instances.
The whole framework is enhanced by the CoModGAN-ADA generator, which ensures the quality and variety of generated histopathology images.
Moreover, the introduction of instance segmentation guided strategy improves the model's ability to extract instance-level information.

The proposed self-supervised learning method is effective in extracting fine-grained features, which are more helpful in dense-prediction tasks than other pretraining methods.
This is confirmed in 7 transfer learning experiments and 9 semi-supervised learning experiments, where our method significantly outperforms the SOTA methods in most cases.

We also investigate whether our method provides discriminative representation for classification tasks.
Although the heatmap visualization in \autoref{grad_semi} and the semi-supervised (0.25\%) results in \autoref{tab:semi_experiment} indicates that our method might get biased to nucleus-level features, which could be not as effective as global ones obtained by discriminative pretraining, the transfer learning results show that we can adapt to the classification task with high-quality.
Moreover, we also perform the linear evaluation protocol \cite{Kolesnikov2019linear} on Kather.
As shown in \autoref{tab:linear prob}, we observe that our features are more linear separable than randomly initialized ones (p$<$0.05).
However, the classification tasks on Kather described in the paper are only conducted with the TUM and NORM classes.
In order to make a more comprehensive analysis of the pretraining methods, we perform an additional multi-class classification task on Kather.
It is noteworthy that we exclude adipose, background, debris, and mucus here, because these classes are nuclei-free, which can hardly benefit from our nucleus-aware pretraining method.
We report the performance on the remaining 5 classes of Kather in \autoref{tab:kather_morecls}.
The results show that our method is still more effective for differentiating various classes besides TUM and NORM, indicated by the highest Acc over other pretraining methods, although the macro F1 score is slightly lower than the previous SOTA method.
In the future, we will focus on extending our method with more discriminative features which supplement the local ones for its border usage in classification tasks.

\begin{table}[t]
\centering
\caption{\label{tab:linear prob}
Linear probing experiments for classification using transfer learning protocol.}
\begin{tabular}{lcc}
\hline
 & \multicolumn{2}{c}{Kather} \\
 & Acc & F1 \\ \hline
Random & 0.5936±0.0158 & 0.4966±0.0147 \\
\rowcolor{gray} Proposed & \textbf{0.7700}±0.0146 & \textbf{0.7589}±0.0146 \\ \hline
\end{tabular}
\end{table}

\begin{table}[t]
\centering
\caption{\label{tab:kather_morecls}
Classification results for more classes besides TUM and NORM on Kather using transfer learning protocol.}
\begin{tabular}{lcc}
\hline
\multicolumn{1}{c}{Method} & Acc & F1 \\
\hline
Baseline & 0.6827±0.0134 & 0.6666±0.0098 \\
ImageNet & 0.8266±0.0041 & 0.7603±0.0021 \\
MoCo v2 & 0.8489±0.0089 & 0.8255±0.0042 \\
Simsiam & 0.7666±0.0030 & 0.7376±0.0029 \\
Mormont & 0.8479±0.0102 & 0.8235±0.0057 \\
DiRA & 0.8412±0.0057 & \textbf{0.8270}±0.0021 \\
Ciga & 0.8390±0.0083 & 0.8225±0.0063 \\
\rowcolor{gray} Proposed & \textbf{0.8511}±0.0073 & 0.8234±0.0024 \\
\hline
\end{tabular}
\end{table}

It is also inspiring to find that the proposed pretraining method is robust to various situations.
First of all, our method is robust to different tissues for pretraining.
We validate this by pretraining on different tissue types and different tumor types.
We collect equal amounts of histopathology images from breast tissues other than colorectal histopathology images for pretraining.
As shown in \autoref{tab:pretrained source}, we observe insignificant differences between the models pretrained on colon or breast (p$=$0.42, AJI).  
We also pretrain on both the colon and breast tissues to assess the potential benefits of leveraging more diverse pretraining sources for our method. 
In this case, we double the training iterations, ensuring that the number of times each image was presented to the network remains unchanged.
It is inspiring to find that the performance on Kumar (AJI) and Kather is slightly improved with the incorporation of such diverse datasets.
The observed improvement can be explained in \autoref{tab:kumar_results_ablation}, which shows that a more diverse pretraining source can mitigate biases towards specific organs.
For example, when only using colon tissues for pretraining, the performance on the stomach, liver, and breast slightly decreased compared to pretraining on breast tissues alone. However, this bias was alleviated when employing the combined dataset, resulting in performance on these organs that match those achieved through pretraining on breast tissues.
Similar results can also be found on the Pannuke dataset reported in \autoref{tab:pannuke_results_ablation}, where the decreased metrics on adrenal, breast, cervix, liver, pancreatic, and uterus recover after adding breast tissues for pretraining.
Therefore,  more stable results may be achieved when we collect more diverse datasets for pretraining.

\begin{table}[t]
\centering
\caption{\label{tab:pretrained source}
Different pretrained sources using transfer learning protocol.}
\resizebox{\columnwidth}{!}{%
\begin{tabular}{lcccc}
\hline
\multirow{2}{*}{Organ} & \multicolumn{2}{c}{Kumar} & \multicolumn{2}{c}{Kather} \\
\multicolumn{1}{c}{} & AJI & F1 & Acc & F1 \\ \hline
Breast & 0.5928±0.0033 & 0.7547±0.0051 & 0.9554±0.0039 & 0.9526±0.0052 \\
\rowcolor{gray} Colon & 0.5932±0.0025 & \textbf{0.7591}±0.0044 & 0.9588±0.0028 & 0.9560±0.0041 \\ 
Breast+Colon & \textbf{0.5961}±0.0052 & 0.7582±0.0061 & \textbf{0.9611}±0.0049 & \textbf{0.9584}±0.0062 \\
\hline
\end{tabular}
}
\end{table}

\begin{table}[t]
\centering
\caption{\label{tab:kumar_results_ablation}
Average AJI across 7 organ types on the Kumar dataset with the transfer learning protocol and different pretrained sources.}
\begin{tabular}{lc>{\columncolor{gray}}cc}
\hline
\multicolumn{1}{c}{Organ} & Breast & Colon & Breast+Colon \\ \hline
Stomach & 0.6275 & 0.6206 & \textbf{0.6278} \\
Colon & 0.4895 & \textbf{0.5008} & 0.5006 \\
Bladder & 0.6270 & \textbf{0.6404} & 0.6338 \\
Prostate & 0.6352 & 0.6302 & \textbf{0.6407} \\
Liver & \textbf{0.5495} & 0.5323 & 0.5451 \\
Kidney & 0.6056 & \textbf{0.6226} & 0.6223 \\
Breast & \textbf{0.6157} & 0.6056 & 0.6087 \\
\hline
Overall & \begin{tabular}[c]{@{}c@{}}0.5928\\ ±0.0033\end{tabular} & \begin{tabular}[c]{@{}c@{}}0.5932\\ ±0.0025\end{tabular} & \begin{tabular}[c]{@{}c@{}}\textbf{0.5970}\\ ±0.0052\end{tabular} \\
\hline
\end{tabular}
\end{table}

\begin{table}[t]
\centering
\caption{\label{tab:pannuke_results_ablation}
Average mPQ+ across 19 tissue types on the Pannuke dataset with the transfer learning protocol and different pretrained sources.}
\begin{tabular}{lc>{\columncolor{gray}}cc}
\hline
\multicolumn{1}{c}{Organ} & Breast & Colon & Breast+Colon \\ \hline
Adrenal & \textbf{0.4061} & 0.3807 & 0.3978 \\
Bile Duct & 0.3673 & 0.3613 & \textbf{0.3813} \\
Bladder & 0.3464 & \textbf{0.3566} & 0.3470 \\
Breast & \textbf{0.4203} & 0.4190 & 0.4201 \\
Cervix & \textbf{0.3711} & 0.3344 & 0.3567 \\
Colon & 0.3704 & \textbf{0.3932} & 0.3836 \\
Esophagus & 0.3998 & 0.3977 & \textbf{0.4087} \\
H\&N* & 0.3698 & 0.3548 & \textbf{0.3866} \\
Kidney & 0.2377 & 0.2335 & \textbf{0.2412} \\
Liver & \textbf{0.4296} & 0.4242 & 0.4279 \\
Lung & 0.2662 & \textbf{0.2886} & 0.2790 \\
Ovarian & 0.4080 & \textbf{0.4278} & 0.4032 \\
Pancreatic & \textbf{0.2840} & 0.2530 & 0.2576 \\
Prostate & 0.3108 & 0.3039 & \textbf{0.3336} \\
Skin & 0.2819 & 0.3070 & \textbf{0.3284} \\
Stomach & 0.3523 & 0.3506 & \textbf{0.3538} \\
Testis & 0.3692 & \textbf{0.4051} & 0.3811 \\
Thyroid & \textbf{0.3374} & 0.3312 & 0.3203 \\
Uterus & \textbf{0.2677} & 0.2578 & 0.2671 \\
\hline
Overall & \begin{tabular}[c]{@{}c@{}}0.4401\\ ±0.0050\end{tabular} & \begin{tabular}[c]{@{}c@{}}0.4432\\ ±0.0022\end{tabular} & \begin{tabular}[c]{@{}c@{}}\textbf{0.4459}\\ ±0.0065\end{tabular} \\
\hline
\end{tabular}
\end{table}

\begin{figure}[t]
	\centering
	\includegraphics[width=.9\columnwidth]{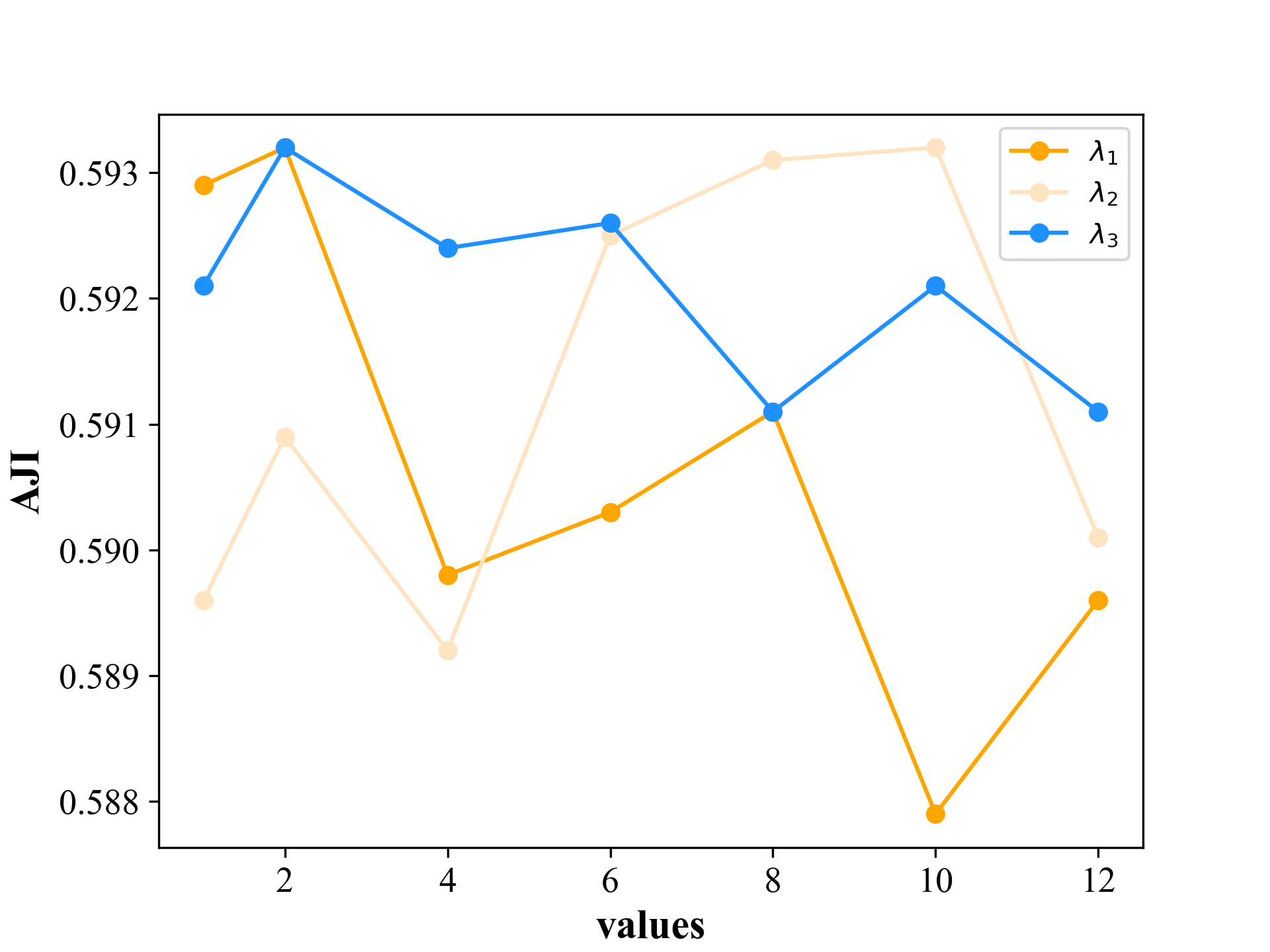}
\caption{Ablation study on the weight-balancing hyperparameters. The experiment is performed on Kumar with transfer learning protocol.}
	\label{ablation_lambda}
\end{figure}

Furthermore, we collect histopathology images scanned from poorly differentiated colorectal tumors, which have less obvious glandular patterns.
It is expected that the glandular distribution of nuclei in the pseudo mask performs worst than the random distribution in such a case. 
However, as reported in \autoref{tab:poor_diff}, we observe comparable results of our method and the baseline (p=0.35), which indicate that our framework can robustly handle the distribution gap between the histopathology images and the mask images.
Secondly, our method is robust to various downstream tasks, which can be seen in \autoref{tab:transfer_experiment} where our method is effective in classification, instance segmentation, semantic segmentation, and multiple instance learning tasks.
Thirdly, our method benefits various architectures for downstream tasks.
We experiment on Panoptic FPN, Mask-RCNN, Hover-Net, UperNet, and C2C for various tasks, and report new SOTA results on Kumar using Hover-Net.
Furthermore, our method is robust to the selection of hyperparameters.
We conducted an additional ablation study to examine the sensitivity of our method to the selection of key hyperparameters. Specifically, we evaluated the impact of changing a weight-balancing hyperparameter while keeping the other hyperparameters fixed. 
The results are presented in \autoref{ablation_lambda}, which reveal that the performance remains relatively stable across various weight-balancing hyperparameter settings within a specific range.

\begin{table}[t]
\centering
\caption{\label{tab:poor_diff}
The performance of our method on poorly differentiated tissues using transfer learning protocol.}
\resizebox{\columnwidth}{!}{%
\begin{tabular}{lcccc}
\hline
\multirow{2}{*}{\begin{tabular}[c]{@{}l@{}}Mask Image\\ Design\end{tabular}} & \multicolumn{2}{c}{Kumar} & \multicolumn{2}{c}{Kather} \\
\multicolumn{1}{c}{} & AJI & F1 & Acc & F1 \\ \hline
Baseline & \textbf{0.5929}±0.0054 & \textbf{0.7546}±0.0067 & \textbf{0.9539}±0.0048 & \textbf{0.9514}±0.0054 \\
\rowcolor{gray} +Distribution & 0.5920±0.0042 & 0.7537±0.0050 & 0.9501±0.0046 & 0.9485±0.0052 \\ 
\hline
\end{tabular}
}
\end{table}

Although the proposed self-supervised pretraining approach has demonstrated promising results for various tasks, some extensions remain to be made.
First of all, the mask images are fixed and limited by the hand-crafted design, where we randomly redistribute parts of the nuclei to match the glandular structures. 
However, a better solution is to make the nuclear distribution vary with different datasets for pertaining, so that the aligned data distribution and the possibility of adopting more diverse datasets for pretraining may lead to better performance.
Strategies such as pseudo-labeling or using pretrained nuclei segmenters may be useful, which will be investigated in the future.
Moreover, the proposed method only focuses on the extraction of local features, which is orthogonal with pretraining approaches that obtain global representation.
It is desirable to research for approaches to embed these methods.
We hope that our method can serve as a baseline for nucleus-aware self-supervised pretraining methods.

\bibliographystyle{IEEEtran}
\bibliography{tmi}

\end{document}